\documentclass[10pt,twocolumn,letterpaper]{article}

\usepackage[pagenumbers]{cvpr}
\usepackage{cuted}
\usepackage{capt-of}
\usepackage{amsthm}
\newtheorem{proposition}{Proposition}

\usepackage{amsmath,amssymb,mathtools}
\usepackage{booktabs}
\usepackage{multirow}
\usepackage{tabularx}
\usepackage{makecell}
\usepackage{bm}
\usepackage{xspace}
\usepackage{graphicx}
\usepackage{adjustbox}
\usepackage{pifont}
\usepackage{placeins}
\usepackage{colortbl}
\makeatletter
\expandafter\gdef\csname ver@everyshi.sty\endcsname{2001/05/15 v3.00}
\makeatother
\usepackage{pgfplots}
\pgfplotsset{compat=1.16}

\definecolor{tblheader}{RGB}{238,240,245}
\definecolor{tblsubheader}{RGB}{247,248,251}
\definecolor{tblband}{RGB}{242,238,248}
\definecolor{tblours}{RGB}{232,236,250}
\definecolor{tblzebra}{RGB}{250,250,252}
\definecolor{tblaccent}{RGB}{76,96,170}

\AtBeginDocument{%
  \rmfamily
  \DeclareFontShape{OT1}{ptm}{m}{scit}{<->ptmrc7t}{}%
  \DeclareFontShape{OT1}{ptm}{b}{scit}{<->ptmbc7t}{}%
}

\newcommand{\nas}{\mathrm{NAS}}

\newcommand{\method}{RELACT}
\usepackage{colortbl}
\newcommand{\ourrow}{\rowcolor{blue!10}}

\definecolor{cvprblue}{rgb}{0.21,0.49,0.74}
\usepackage[pagebackref,breaklinks,colorlinks,allcolors=cvprblue]{hyperref}

\def\confName{CVPR}
\def\confYear{2027}

\title{Relation Before Action: Explicit Relation Modeling for Latent Action Pretraining}

\author{%
Boning Zhang$^{1,2}$ \qquad Junyu Gao$^{1,2}$ \qquad Changsheng Xu$^{1,2,3}$\textsuperscript{*}\\[4pt]
{\small $^{1}$State Key Laboratory of Multimodal Artificial Intelligence Systems (MAIS),}\\
{\small Institute of Automation, Chinese Academy of Sciences (CASIA)}\\
{\small $^{2}$School of Artificial Intelligence, University of Chinese Academy of Sciences (UCAS)}\\
{\small $^{3}$Peng Cheng Laboratory, Shenzhen, China}\\
{\small \texttt{zhangboning2025@ia.ac.cn} \quad \texttt{junyu.gao@nlpr.ia.ac.cn} \quad \texttt{csxu@nlpr.ia.ac.cn}}\\
{\small \textsuperscript{*}Corresponding author.}%
}

\begin{document}
\maketitle

\begin{strip}
    \centering
    \includegraphics[width=0.98\textwidth]{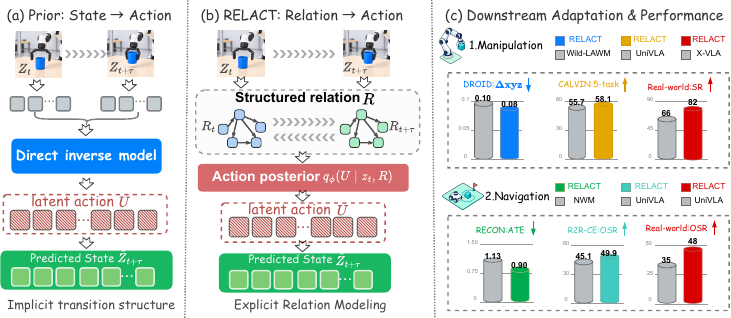}
    \captionof{figure}{
    \textbf{Relation before action.}
    Conventional latent-action methods infer actions directly from paired observations, leaving the underlying transition relation implicit.
    In contrast, \method{} explicitly constructs a structured relation representation $R$ before abstracting it into a compact, state-conditioned latent action $U$.
    Extensive experiments across navigation and manipulation demonstrate consistent improvements in predictive planning and policy adaptation, highlighting the generality and transferability of relation-first action representations.
    }
    \label{fig:teaser}
\end{strip}

\begin{abstract}
    Learning latent actions from unlabeled video sequences offers a promising way to acquire reusable action representations without relying on embodiment-specific control annotations. However, existing latent-action pretraining approaches commonly couple the modeling of visual transitions with action inference, directly mapping paired observations or derived transition cues into latent action codes. This formulation leaves two distinct tasks entangled: characterizing the relational structure of an observed transition and abstracting it into a compact action representation.
    We propose \method{}, a relation-first framework that explicitly separates transition relation modeling from predictive action abstraction. \method{} first constructs a structured relation representation \(R\) that captures how visual elements and their interactions evolve across observations. It then infers a compact, state-conditioned stochastic latent action \(U\), whose predictive effects are learned through action-conditioned latent dynamics. The resulting action representation can be grounded in native control spaces for model-based planning or used to supervise downstream embodied policies.
    Extensive experiments across in-domain and out-of-domain image-goal navigation, goal-conditioned manipulation, and downstream policy adaptation demonstrate improved predictive action representations and consistent gains in planning and control, supporting the effectiveness and generality of relation-first latent action pretraining.
\end{abstract}

\section{Introduction}
\label{sec:intro}

World action models (WAMs) learn predictive dynamics for embodied agents, but their reliance on native action spaces often limits transfer across embodiments and control interfaces.
To overcome this limitation, recent works explore \emph{latent action pretraining}, which learns action representations from largely unlabeled visual experience and subsequently grounds them in native control spaces with limited action supervision~\cite{edwards2019imitating,bruce2024genie,ye2025lapa,bu2025univla,chen2026villax,garrido2026wildlawm}.
A common formulation directly infers latent actions from paired visual observations or their feature differences, and uses the inferred actions to predict future states.

However, these formulations leave an important distinction implicit: \emph{what changes in a scene is not necessarily what a latent action needs to encode.}
Paired visual representations can implicitly encode temporal relationships, but existing approaches typically compress such information directly into latent action codes without explicitly organizing the relational structure of the transition.
This couples two conceptually distinct tasks: identifying how scene elements and their interactions change, and determining which aspects of those changes are relevant to action-conditioned dynamics.
As a result, the action encoder must simultaneously discover transition structure and abstract its predictive effects.
Under limited representation capacity, this entanglement may encourage latent actions to retain incidental future-state information rather than selectively capturing transition-relevant effects.
Although bottlenecks, regularization, and auxiliary objectives can mitigate this issue, they do not explicitly separate transition representation from action abstraction.

This distinction motivates us to rethink how transition information should be represented before action inference.
Prior research on relational inductive biases suggests that explicitly modeling interactions among scene elements can facilitate structured dynamics learning~\cite{battaglia2018relational,kipf2020contrastive}.
This perspective suggests that the organization of transition evidence itself can serve as a useful inductive bias for latent-action learning.
Complementarily, the information bottleneck perspective motivates learning compact representations that preserve predictive information~\cite{alemi2017deep}.
Inspired by these complementary perspectives, we propose a simple principle: \emph{model the relation before the action}.

We introduce \method{}, a relation-first latent-action pretraining framework that explicitly separates these two roles.
We define a \emph{transition relation} as the change in directed pairwise interactions among softly corresponding latent scene elements across observations.
Specifically, a frozen paired-view visual encoder provides jointly contextualized features, from which shared scene queries extract comparable latent elements at the two time steps.
Changes in their directed interactions are aggregated into a structured relation representation $R$, without requiring explicit object identities or ground-truth tracking.
A state-conditioned stochastic posterior then infers a compact latent action $U$ from $(Z_t,R)$, where the current state provides context for interpreting the observed relational change.
The resulting action conditions a latent dynamics model to predict future visual representations.

This relation-first factorization offers two complementary advantages.
First, explicitly representing changes in inter-element interactions introduces structural bias into action inference, rather than leaving the discovery of transition structure entirely to a direct inverse model.
Second, separating relation modeling from action abstraction allows the two representations to serve different purposes: $R$ organizes evidence of the observed transition, whereas $U$ compresses that evidence into a state-conditioned representation of its predictive effects.
Importantly, the forward dynamics model receives only the visual history and $U$, not the explicit relation or future observation.
Combined with predictive supervision, action-specific objectives, and capacity regularization, this design encourages the learned action to retain transition-relevant information without requiring it to reconstruct every detail of the observed relation.

After pretraining, the learned action representation can be reused through lightweight, embodiment-specific interfaces while keeping the pretrained representation and dynamics fixed.
For model-based planning, a state-conditioned grounding interface maps native controls into the latent action space, enabling candidate actions to be evaluated through predictive rollouts.
For downstream policy adaptation, a frozen \method{} teacher provides latent transition targets that supervise a lightweight residual adapter, without requiring future observations at inference time.
These two forms of reuse allow the same relation-first learning principle to support both predictive planning and policy refinement.

We evaluate \method{} across navigation and manipulation, covering in-domain planning, out-of-domain transfer, downstream policy adaptation, and real-world robotic deployment.
Across these settings, \method{} produces predictive and transition-specific latent action representations that improve planning and control.
In particular, controlled comparisons with alternative transition representations demonstrate the contribution of explicit relational modeling beyond access to pretrained paired-view features.

Our contributions are threefold:
\begin{itemize}
    \item We identify the coupling between transition description and action abstraction in latent-action pretraining, and introduce a \textbf{relation-first formulation} that explicitly separates these two roles.

    \item We propose \textbf{\method{}}, which models correspondence-aware changes in directed interactions among latent scene elements before abstracting them into compact, state-conditioned stochastic actions for predictive dynamics.

    \item We demonstrate the effectiveness of \textbf{\method{} across navigation and manipulation}, including planning, domain transfer, and downstream policy adaptation. Controlled ablations establish the practical benefit of explicit relation modeling over alternative transition representations.
\end{itemize}

\section{Related Work}
\label{sec:related}

\noindent\textbf{Latent-action learning and world models.}
Latent-action learning infers action-like representations from visual experience for future prediction, policy pretraining, and control~\cite{edwards2019imitating,bruce2024genie}.
LAPA~\cite{ye2025lapa} learns discrete latent actions from videos, while UniVLA~\cite{bu2025univla} and villa-X~\cite{chen2026villax} extend latent-action pretraining toward task-aware manipulation.
LAWM~\cite{tharwat2025lawm}, Wild-LAWM~\cite{garrido2026wildlawm}, SCAR~\cite{liu2026scar}, and LAC-WM~\cite{huang2026lacwm} explore predictive latent-action dynamics, while Motus~\cite{bi2026motus} and CycleMimic~\cite{chen2026cyclemimic} investigate motion-based representations and action consistency, respectively.
These approaches can capture transition information, but generally do not explicitly organize changes in inter-element relations before action inference.
Closely related, FLAM~\cite{wang2026flam} factorizes scene dynamics into action-bearing factors with separate latent actions, whereas \method{} models changes in directed inter-element relations before abstracting them into a compact, state-conditioned latent action.

\noindent\textbf{Relational dynamics and action grounding.}
Relational inductive biases provide a foundation for modeling interactions among scene elements~\cite{battaglia2018relational}, while structured world models exploit such interactions for learning predictive dynamics~\cite{kipf2020contrastive}.
These works motivate structured transition representations, but do not directly address their role as an intermediate stage in latent-action inference.
Meanwhile, latent-action methods such as LAPA~\cite{ye2025lapa}, Wild-LAWM~\cite{garrido2026wildlawm}, and LAC-WM~\cite{huang2026lacwm} connect learned action spaces to native controls for downstream use.
\method{} connects these directions by explicitly modeling changes in directed pairwise interactions before learning a compact predictive action representation.
After domain-specific pretraining, the learned representation is reused through lightweight interfaces for native-action planning or downstream policy adaptation.

\section{Method}
\label{sec:method}

\subsection{Overview}
\label{sec:formulation}

Let $I_t$ and $I_{t+\tau}$ denote two visual observations separated by a temporal horizon $\tau$.
Our goal is to learn a compact latent action $U_{t:t+\tau}$ that explains the observed transition and supports predictive dynamics and downstream control.

An observed transition and the action underlying it are distinct.
The former describes how observations change, whereas the latter characterizes the dynamical effect of that change under the current state.
Existing latent-action methods often infer actions directly from paired observations, leaving the representation of the transition implicit within a single inverse model.
We instead introduce an explicit intermediate relation and formulate latent-action learning as a \emph{relation-first predictive abstraction}:
\begin{equation}
\begin{aligned}
Z_t &= E(I_t), \quad
Z_{t+\tau}=E(I_{t+\tau}),\\
V_{t,t+\tau}
&=\Psi_{\Omega}(I_t,I_{t+\tau}),\\
R_{t,t+\tau}
&=\mathcal R_{\eta}(V_{t,t+\tau}),\\
U_{t:t+\tau}
&\sim q_\phi(U\mid Z_t,R_{t,t+\tau}),\\
\widehat Z_{t+\tau}
&=F_\theta(\mathcal H_t,U_{t:t+\tau}).
\end{aligned}
\label{eq:factorization}
\end{equation}

\begin{figure*}[t]
    \centering
    \includegraphics[width=0.99\textwidth]{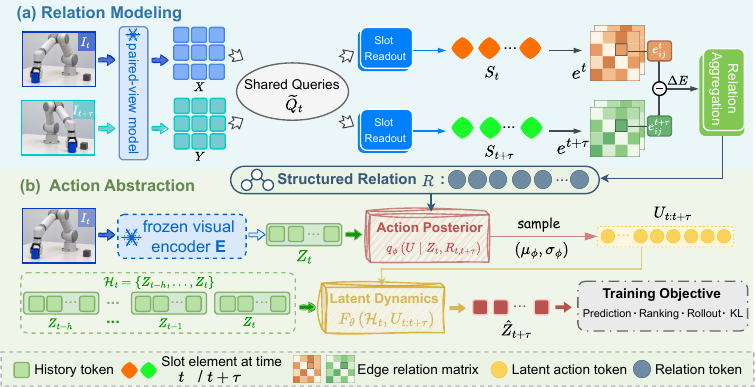}
    \caption{
    \textbf{\method{}: Relation Before Action.}
    (a) \emph{Transition relation modeling:} A frozen paired-view encoder provides contextualized tokens for the observation pair. Shared queries read out comparable latent scene elements from both observations, and changes in their directed relations are aggregated into a structured relation representation $R$.
    (b) \emph{Predictive action abstraction:} A state-conditioned posterior infers a compact latent action $U$ from the current visual state $Z_t$ and the relation $R$. This action conditions the dynamics to predict future visual representations. Predictive and action-specific objectives jointly supervise the relation-to-action pathway and latent dynamics during pretraining.
    }
    \label{fig:method}
\end{figure*}

Here, $E$ encodes visual states, $\Psi_{\Omega}$ is a frozen paired-view encoder, and $\mathcal R_{\eta}$ organizes its jointly contextualized evidence into a structured transition relation $R$.
A state-conditioned posterior $q_\phi$ then abstracts the relation into a latent action $U$, whose consequence is predicted by $F_\theta$ given the visual history $\mathcal H_t$.
The overall framework is illustrated in Fig.~\ref{fig:method}.

The two representations serve different purposes.
The relation represents evidence of the observed transition, whereas the latent action provides a compact representation of its predictive effect.
This separation allows transition information to be represented before determining which aspects are relevant to action-level dynamics.
Sec.~\ref{sec:relation} models the observed transition by extracting comparable scene elements from the paired observations and aggregating changes in their directed relations.
Sec.~\ref{sec:action} abstracts that relation into the latent action $U$ and gives the predictive objective used to learn it.
Sec.~\ref{sec:reuse} reuses the learned action space for model-based planning and downstream policy adaptation.

\subsection{Transition Relation Modeling}
\label{sec:relation}

We instantiate the relation-modeling stage of our relation-first formulation through a structured pipeline that combines shared scene queries, directed inter-element interactions, and aggregation of their changes across observations.
As illustrated in Fig.~\ref{fig:method}(a), shared queries first extract comparable latent scene elements from paired-view features, after which changes in their directed pairwise relations are aggregated into a structured transition representation $R$.

\noindent\textbf{Shared scene queries and slot readout.}\ 
Given an observation pair $(I_t,I_{t+\tau})$, a frozen paired-view encoder produces jointly contextualized tokens
\begin{equation*}
(X,Y)=\Psi_{\Omega}(I_t,I_{t+\tau}),
\qquad X,Y\in\mathbb{R}^{N\times d}.
\end{equation*}
Starting from spatially anchored scene queries, we condition the queries on the source-view features and reuse the resulting query set for both observations, establishing a shared basis for relational comparison:
\begin{equation}
\begin{aligned}
S_t &= \operatorname{Readout}(\widetilde Q_t, X),\\
S_{t+\tau} &= \operatorname{Readout}(\widetilde Q_t, Y),
\end{aligned}
\label{eq:relation_slots}
\end{equation}
where $S_t,S_{t+\tau}\in\mathbb{R}^{K_s\times d_s}$ denote $K_s$ slot-like scene elements. The spatial anchors keep these queries from reading the same image region. The correspondence is soft and does not assume object identities or explicit tracking.

\noindent\textbf{Directed relational change and aggregation.}\ 
Each ordered pair of elements carries an edge that combines their feature difference, their image-plane offset, and an asymmetric bilinear interaction:
\begin{equation}
\begin{aligned}
e_{ij}^{\nu}
={}&
W_d(s_j^{\nu}-s_i^{\nu})
+W_p(p_j^{\nu}-p_i^{\nu})\\
&+\kappa W_b\!\left[(W_q\bar s_i^{\nu})\odot(W_k\bar s_j^{\nu})\right],
\end{aligned}
\label{eq:relation_edge}
\end{equation}
where $\bar s_i^{\nu}=\operatorname{LN}(s_i^{\nu})$, $\odot$ is element-wise multiplication, and every map is learned. The node set also contains one node fixed at the image center. When the scene elements undergo a common image-plane displacement, offsets among the elements can stay nearly unchanged, while offsets to this fixed node record that shared motion. The transition is the change of these edges,
\begin{equation}
\begin{aligned}
\Delta e_{ij} &= e_{ij}^{t+\tau}-e_{ij}^{t},\\
R_{t,t+\tau} &= \mathcal A_{\eta}\!\left(\{\Delta e_{ij}\}_{i\neq j}\right),
\end{aligned}
\label{eq:relation}
\end{equation}
with $R_{t,t+\tau}\in\mathbb{R}^{K_R\times d_R}$. The aggregator uses source-frame spatial offsets to weight incoming and outgoing relational changes, preserving their directional roles in the resulting relation tokens. Appendix~\ref{app:relation_encoder} specifies the query conditioning, widths, and aggregation weights.

\subsection{Predictive Action Abstraction}
\label{sec:action}

The structured relation $R$ characterizes how scene elements and their directed interactions change across observations. However, a latent action need not preserve every detail of this relational transition. We therefore abstract $R$ into a compact, state-conditioned action $U$, encouraging it to retain the information most relevant to predictive dynamics. This abstraction is learned jointly with the latent dynamics model through the objectives below.

\noindent\textbf{State-conditioned action abstraction.}\ 
The same relational change can have different predictive implications under different originating states. We therefore infer a latent action from both the current visual state and the structured relation. Cross-attention integrates relation tokens with state features to parameterize a diagonal-Gaussian posterior:
\begin{equation}
q_\phi(U\mid Z_t,R)
=\mathcal N\!\left(\mu_\phi,\operatorname{diag}\sigma_\phi^2\right),
\label{eq:posterior}
\end{equation}
where the mean and variance depend on the current state and relation. The sampled action then conditions a history-aware latent dynamics model through a residual prediction:
\begin{equation}
\widehat Z_{t+\tau}
=F_\theta(\mathcal H_t,U)
=Z_t+\Delta_\theta(\mathcal H_t,U).
\label{eq:fdm}
\end{equation}
Here, $\mathcal H_t$ contains up to $h$ preceding visual states. Crucially, the dynamics model receives neither the explicit relation nor the future target; transition evidence can affect its prediction only through the compact action variable. This implements predictive action abstraction without requiring the action to reconstruct every detail of the relation (Appendix~\ref{app:predictive_theory}).

\noindent\textbf{Predictive and action-specific objectives.}\ 
We train the relation encoder, action posterior, and latent dynamics jointly with complementary predictive and action-specific objectives.
First, future-feature prediction combines token-averaged cosine distance and feature MSE against a stop-gradient target, encouraging the action channel to retain information needed to explain the observed transition.
Second, because accurate prediction alone may permit the dynamics to ignore its action input, we compare matched and mismatched latent actions under the same visual history.
Writing $d_i(U)$ for the cosine prediction error on transition $i$ when conditioned on action $U$, the action-ranking loss is
\begin{equation}
\mathcal L_{\mathrm{rank}}
=\mathbb E_{i,j\ne i}
\big[m+d_i(U_i)-d_i(U_j)\big]_+.
\label{eq:rank}
\end{equation}
This encourages an action inferred from a transition to explain that transition better than an action drawn from another example.
Finally, a KL penalty to a unit Gaussian constrains the posterior's information capacity, while short detached rollouts promote multi-step consistency.

Together, these terms form the pretraining objective
\begin{equation}
\mathcal L_{\mathrm{pre}}
=\mathcal L_{\mathrm{pred}}
+\gamma\mathcal L_{\mathrm{rank}}
+\beta\mathcal L_{\mathrm{KL}}
+\lambda_{\mathrm{roll}}\mathcal L_{\mathrm{roll}}.
\label{eq:total}
\end{equation}
The prediction and capacity terms promote a compact predictive abstraction; action ranking and rollout further encourage transition-specific action usage and temporal consistency.

\subsection{Reusing the Learned Action Space}
\label{sec:reuse}

\begin{figure}[t]
    \centering
    \includegraphics[width=\columnwidth]{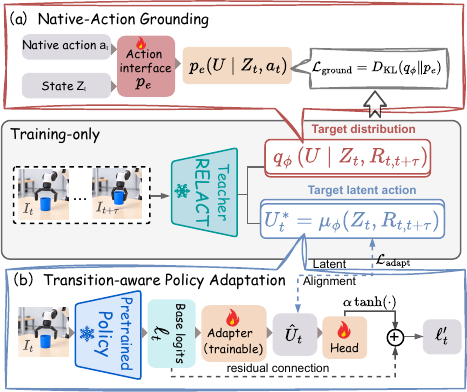}
    \caption{
    \textbf{Reusing the learned action space.}
    After pretraining, the learned latent action space is reused in two complementary ways.
    (a) Native controls are grounded into the latent action space for predictive planning with frozen latent dynamics.
    (b) A frozen \method{} teacher supervises a lightweight adapter that refines an existing policy through latent transition targets.
    Training-only teacher paths are removed at inference.
    }
    \label{fig:reuse}
\end{figure}

After pretraining, the learned action representation can be reused through lightweight, embodiment-specific interfaces while keeping the pretrained relation-to-action model fixed.
We consider two complementary forms of reuse, illustrated in Fig.~\ref{fig:reuse}.

\noindent\textbf{Native-action grounding for planning.}\ 
Given labeled transitions $(I_t,a_t^e,I_{t+\tau})$ from embodiment $e$, the frozen relation-to-action posterior serves as a teacher for a state-conditioned Gaussian interface $p_e(U\mid Z_t,a_t^e)$. We align native controls with their observed latent effects by minimizing
\begin{equation}
\mathcal L_{\mathrm{ground}}
=D_{\mathrm{KL}}\!\left(
q_\phi(U\mid Z_t,R_{t,t+\tau})
\,\Vert\,
p_e(U\mid Z_t,a_t^e)
\right).
\label{eq:grounding}
\end{equation}
This interface maps an embodiment's control into the learned action space without modifying the pretrained posterior or dynamics. At planning time, candidate native-action sequences are mapped into latent actions and evaluated through rollouts of the frozen $F_\theta$. Their predicted goal costs determine which native controls to execute. The grounding interface uses $Z_t$, whereas the dynamics model may additionally use temporal context $\mathcal H_t$. Further planning details are provided in Appendix~\ref{app:grounding}.

\noindent\textbf{Transition-aware policy adaptation.}\ 
The action space can also provide supervision to an existing embodied policy without online world-model rollout. Given a frozen policy representation $h_t$, a lightweight adapter predicts $\widehat U_t=A_\psi(h_t)$ and refines the original action logits through a bounded residual:
\begin{equation}
\ell_t'
=
\ell_t+\alpha_{\mathrm{res}}\tanh(G_\psi(\widehat U_t)).
\label{eq:policy_residual}
\end{equation}
During training, expert observation pairs provide teacher targets $U_t^\star=\mu_\phi(Z_t,R_{t,t+\tau})$, while the adapter predicts them from the policy features available at inference. Training combines the original task objective, latent-action alignment, and conservative residual regularization; the residual output layer is zero-initialized to preserve the base policy initially. At inference, the adapter requires neither future observations nor the frozen teacher. Further details are provided in Appendix~\ref{app:policy_adapter}.

Both interfaces reuse $U$ as a predictive connection between observed transitions and embodied control, either through model-based evaluation of native controls or through lightweight policy refinement.

\section{Experiments}
\label{sec:experiments}

\subsection{Experimental Setup}
\label{sec:setup}

\noindent\textbf{Tasks and datasets.}\ 
We evaluate \method{} along two forms of action-space reuse introduced in
Sec.~\ref{sec:reuse}.
For model-based planning, we consider image-goal navigation and
goal-conditioned manipulation.
Navigation pretraining uses
RECON~\cite{shah2021recon},
HuRoN~\cite{hirose2023huron},
TartanDrive~\cite{triest2022tartandrive},
and SCAND~\cite{karnan2022scand};
Go Stanford~\cite{hirose2018gonet} is excluded from world-model training
and used only with a scalar waypoint-scale calibration fit on its training split.
For out-of-domain transfer, the world model is pretrained on
HM3D~\cite{ramakrishnan2021hm3d},
VLNVerse~\cite{lin2025vlnverse}, and HSSD~\cite{khanna2024hssd},
and transferred to RECON without updating the pretrained world-model weights.
For goal-conditioned manipulation planning, we follow the DROID~\cite{khazatsky2024droid}
protocol of Wild-LAWM~\cite{garrido2026wildlawm}.

For policy-based reuse, we instantiate the transition-aware adapter of
Sec.~\ref{sec:reuse} on a pretrained UniVLA~\cite{bu2025univla} policy and evaluate it on
R2R-CE~\cite{anderson2018vision,krantz2020beyond,savva2019habitat} and CALVIN~\cite{mees2022calvin}.
Finally, we evaluate the resulting navigation and manipulation systems on
physical robots.
Dataset splits, action interfaces, metric definitions, and the planning protocol are in Appendix~\ref{app:eval}.

\noindent\textbf{Baselines and metrics.}\ 
Image-goal navigation is compared with
GNM~\cite{shah2023gnm},
NoMaD~\cite{sridhar2024nomad},
NWM~\cite{bar2025nwm},
and CoME~\cite{stapf2026come},
and is scored by ATE and RPE.
Out-of-domain RECON transfer uses the same two errors against Wild-LAWM~\cite{garrido2026wildlawm}.
DROID planning is compared with
V-JEPA~2-AC~\cite{assran2025vjepa2},
V-JEPA~2 + WM~\cite{terver2026physicalplanning},
and the Sparse, Discrete, and Noisy variants of Wild-LAWM~\cite{garrido2026wildlawm},
and is scored by $\Delta xyz$.
Policy adaptation compares the frozen UniVLA policy, a capacity-matched residual adapter, and the same adapter with \method{} supervision.
R2R-CE reports oracle success.
CALVIN reports 5-task chain success and average completed length.
Real-robot trials report task success.
Definitions of these metrics are in Appendix~\ref{app:eval}.

\noindent\textbf{Implementation.}\ 
$E$ is a frozen V-JEPA~2.1 ViT-B/16~\cite{murlabadia2026vjepa21,assran2025vjepa2,dosovitskiy2021vit} on $384\times384$ frames, and $\Omega$ is frozen {VGGT-$\Omega$}~\cite{wang2026vggtomega} on $512\times512$ inputs.
$R$ follows Sec.~\ref{sec:relation}: shared queries read out comparable scene elements, and changes in their directed relations are aggregated into the transition relation.
$U$ is a 128-dimensional Gaussian, and $F_\theta$ conditions on $h=4$ history frames with detached closed-loop rollouts of $L_{\mathrm{roll}}=4$.
Training uses AdamW with learning rate $3\times10^{-4}$, weight decay $0.01$, and global batch 8.
In-domain navigation runs for 40{,}000 updates on RECON, HuRoN, TartanDrive, and SCAND, with each source capped at 25{,}000 clips, and one checkpoint is shared across the five navigation domains.
The indoor transfer model runs for 32{,}000 updates on HM3D, VLNVerse, and HSSD.
DROID is trained separately at 4\,fps on Cartesian translations and is not initialized from either navigation checkpoint.
Loss weights, the comparison against the Wild-LAWM budget, and step-time measurements are in Appendix~\ref{app:pretrain}.

\begin{table*}[t!]
\caption{
\textbf{Image-goal navigation across five domains.}
Lower ATE and RPE indicate better performance.
\method{} uses one shared navigation checkpoint across all five domains.
RECON, TartanDrive, and SCAND use the official evaluation splits.
HuRoN is evaluated on our split because the high-resolution trajectories used
by prior work are not publicly available.
Go Stanford is excluded from world-model training; only a scalar waypoint-scale
calibration is fit on its training split.
}
\label{tab:nav_main}
\centering
\small
\setlength{\tabcolsep}{4pt}
\begin{tabular}{@{}l cc cc cc cc cc@{}}
\toprule
& \multicolumn{2}{c}{RECON}
& \multicolumn{2}{c}{HuRoN}
& \multicolumn{2}{c}{TartanDrive}
& \multicolumn{2}{c}{SCAND}
& \multicolumn{2}{c}{Go Stanford} \\
\cmidrule(lr){2-3}
\cmidrule(lr){4-5}
\cmidrule(lr){6-7}
\cmidrule(lr){8-9}
\cmidrule(lr){10-11}
Method
& ATE$\downarrow$ & RPE$\downarrow$
& ATE$\downarrow$ & RPE$\downarrow$
& ATE$\downarrow$ & RPE$\downarrow$
& ATE$\downarrow$ & RPE$\downarrow$
& ATE$\downarrow$ & RPE$\downarrow$ \\
\midrule
GNM~\cite{shah2023gnm}
& 1.87 & 0.73
& 3.07 & 1.03
& 2.21 & 0.72
& 2.12 & 0.61
& -- & -- \\
NoMaD~\cite{sridhar2024nomad}
& 1.95 & 0.53
& 2.73 & 0.72
& 4.51 & 0.95
& 2.24 & 0.49
& -- & -- \\
NWM~\cite{bar2025nwm}
& 1.13 & 0.35
& 5.56 & 1.19
& 5.85 & 1.22
& 1.28 & 0.33
& 2.29 & 0.61 \\
CoME~\cite{stapf2026come}
& 0.96 & \textbf{0.28}
& -- & --
& -- & --
& -- & --
& -- & -- \\
\midrule
\ourrow
\textbf{\method{}}
& \textbf{0.90} & 0.31
& \textbf{1.95} & \textbf{0.61}
& \textbf{1.34} & \textbf{0.39}
& \textbf{0.90} & \textbf{0.24}
& \textbf{2.00} & \textbf{0.52} \\
\bottomrule
\end{tabular}
\end{table*}

\subsection{Model-Based Planning and Transfer}
\label{sec:planning}

\noindent\textbf{In-domain image-goal navigation.}\ 
We first evaluate whether the learned latent action space supports predictive
planning across navigation domains.
Table~\ref{tab:nav_main} reports one shared navigation checkpoint on
RECON, HuRoN, TartanDrive, SCAND, and Go Stanford.
\method{} obtains the lowest ATE on all five reported domains for which
protocol-matched comparisons are available or can be recomputed.
On RECON, it improves ATE from $0.96$ for CoME to $0.90$ while remaining
competitive in RPE.
The same checkpoint also achieves $1.95/0.61$ on HuRoN,
$1.34/0.39$ on TartanDrive, $0.90/0.24$ on SCAND, and
$2.00/0.52$ on Go Stanford, indicating that the learned action representation
supports planning across substantially different visual environments.

\noindent\textbf{Out-of-domain transfer.}\ 
We next test whether relation-first pretraining transfers when the world model
never observes the target navigation datasets during pretraining.
Here, \method{} is pretrained only on HM3D, VLNVerse, and HSSD;
the resulting world model is frozen and only a lightweight RECON interface is
fit on the target training split.
As shown in Table~\ref{tab:transfer_main},
\method{} achieves $1.05$ ATE and $0.34$ RPE, compared with
$1.40$/$0.40$ for the strongest reported Wild-LAWM variant.
Appendix~\ref{app:pretrain} gives the pretraining budgets:
\method{} uses fewer clip-updates than Wild-LAWM and still reaches this lower error.

\noindent\textbf{Goal-conditioned manipulation planning.}\ 
The same relation-first recipe transfers directly to goal-conditioned manipulation.
On DROID, under the Wild-LAWM planning protocol
(Appendix~\ref{app:eval}),
\method{} rolls out in latent space toward the visual goal and never decodes predicted states to RGB.
The resulting error is $\Delta xyz=0.08$\,m, below V-JEPA~2-AC and every reported Wild-LAWM variant.
Together with the navigation results, this shows that relation-to-action modeling remains effective beyond wheeled locomotion.

\begin{table}[t]
\caption{
\textbf{Out-of-domain navigation and manipulation planning.}
Top: RECON navigation with a world model pretrained on separate indoor environments and frozen during target-domain adaptation.
Bottom: goal-conditioned manipulation planning on DROID under the Wild-LAWM evaluation protocol.
Navigation and manipulation use separately pretrained models.
Lower is better for all metrics.
}
\label{tab:transfer_main}
\centering
\small
\setlength{\tabcolsep}{4pt}
\begin{tabular}{@{}llccc@{}}
\toprule
Domain & Method
& ATE$\downarrow$
& RPE$\downarrow$
& $\Delta xyz$$\downarrow$ \\
\midrule
\multirow{4}{*}{\begingroup\setlength{\tabcolsep}{0pt}\makecell[c]{RECON\\OOD}\endgroup}
& Wild-LAWM (Sparse)~\cite{garrido2026wildlawm}
& 1.43 & 0.42 & -- \\
& Wild-LAWM (Discrete)~\cite{garrido2026wildlawm}
& 1.48 & 0.42 & -- \\
& Wild-LAWM (Noisy)~\cite{garrido2026wildlawm}
& 1.40 & 0.40 & -- \\
\ourrow
& \textbf{\method{}}
& \textbf{1.05} & \textbf{0.34} & -- \\
\midrule
\multirow{6}{*}{DROID}
& V-JEPA~2-AC~\cite{assran2025vjepa2}
& -- & -- & 0.15 \\
& V-JEPA~2 + WM~\cite{terver2026physicalplanning}
& -- & -- & \textbf{0.05} \\
& Wild-LAWM (Sparse)~\cite{garrido2026wildlawm}
& -- & -- & 0.13 \\
& Wild-LAWM (Discrete)~\cite{garrido2026wildlawm}
& -- & -- & 0.14 \\
& Wild-LAWM (Noisy)~\cite{garrido2026wildlawm}
& -- & -- & 0.10 \\
\ourrow
& \textbf{\method{}}
& -- & -- & 0.08 \\
\bottomrule
\end{tabular}
\end{table}

\subsection{Transition-Aware Policy Adaptation}
\label{sec:policy_exp}

We next evaluate the second reuse mechanism in Sec.~\ref{sec:reuse}:
instead of performing online rollout through $F_\theta$, the pretrained latent
action space supervises a lightweight adapter attached to an existing embodied
policy.
Our main comparison keeps the downstream policy, adapter capacity, training
data, and optimization fixed, and varies only whether the adapter receives
transition supervision from \method{}.
The adaptation setup and reported metrics are in Appendix~\ref{app:eval}.

\noindent\textbf{Vision-and-language navigation.}\ 
We instantiate the adapter on UniVLA and evaluate on R2R-CE.
The base UniVLA policy directly predicts native navigation actions.
The residual baseline adds the same adapter and residual branch used by our
method but trains them only with the downstream action objective.
\method{} additionally supervises the adapter with the frozen latent action
$U_t^\star$ inferred from expert transitions.
This comparison isolates whether the learned transition representation provides useful supervision beyond the additional adapter capacity.
Table~\ref{tab:r2r_adapt} reports oracle success.
The residual adapter alone raises UniVLA from $45.1$ to $47.5$, and latent
transition supervision further raises it to $49.9$.

\begin{table}[t]
\caption{
\textbf{Transition-aware policy adaptation on R2R-CE.}
Higher is better.
All adapted variants use the same UniVLA backbone and residual adapter.
Only \method{} uses latent transition supervision.
}
\label{tab:r2r_adapt}
\centering
\small
\setlength{\tabcolsep}{4pt}
\begin{tabular}{@{}lc@{}}
\toprule
Method & Oracle success$\uparrow$ \\
\midrule
UniVLA & 45.1 \\
UniVLA + residual adapter & 47.5 \\
\ourrow
\textbf{UniVLA + \method{} supervision} & \textbf{49.9} \\
\bottomrule
\end{tabular}
\end{table}

\noindent\textbf{Language-conditioned manipulation.}\ 
We apply the same adaptation mechanism to CALVIN without changing the
relation-to-action pretraining objective.
As in R2R-CE, the base policy remains frozen and only the lightweight adapter
and residual branch are optimized.
The experiment tests whether transition-space supervision remains useful when
the downstream task changes from navigation to articulated manipulation.
Table~\ref{tab:calvin_adapt} shows that the residual adapter alone does not
improve the frozen policy: 5-task success moves from $55.7$ to $55.5$ and
average length from $3.736$ to $3.702$.
Latent transition supervision raises both to $58.1$ and $3.812$.

\begin{table}[t]
\caption{
\textbf{Transition-aware policy adaptation on CALVIN.}
5-task success is the fraction of sequences that complete five tasks in a row.
Average length is the mean number of completed tasks, with a maximum of 5.
Higher is better for both.
}
\label{tab:calvin_adapt}
\centering
\small
\setlength{\tabcolsep}{4pt}
\begin{tabular}{@{}lcc@{}}
\toprule
Method & 5-task$\uparrow$ & Avg.\ len$\uparrow$ \\
\midrule
UniVLA & 55.7 & 3.736 \\
UniVLA + residual adapter & 55.5 & 3.702 \\
\ourrow
\textbf{UniVLA + \method{} supervision} & \textbf{58.1} & \textbf{3.812} \\
\bottomrule
\end{tabular}
\end{table}

\subsection{Real-World Evaluation}
\label{sec:real_robot}

We deploy pretrained policies on physical robots without updating the pretrained \method{} representation.
Navigation uses UniVLA on a Unitree Go2, and manipulation uses {X-VLA}~\cite{zheng2025x} on a six-axis Piper arm.
Representative trials are shown in Fig.~\ref{fig:real_robot}, and the results are reported in Table~\ref{tab:real_robot}.
Deployment conditions are in Appendix~\ref{app:real}.

\noindent\textbf{Real-robot navigation.}\ 
The Unitree Go2 follows image goals from onboard RGB.
We evaluate both policies over 100 trials in eight scenes, five indoor and three outdoor.
Table~\ref{tab:real_robot} reports oracle success rates of $35\%$ for UniVLA and $48\%$ for UniVLA + \method{}.
At test time the robot executes the frozen UniVLA policy and the trained adapter, with no future observations.

\noindent\textbf{Real-robot manipulation.}\ 
The Piper arm runs {X-VLA}~\cite{zheng2025x} on five tasks, ten trials each.
Table~\ref{tab:real_robot} reports task success rates of $66\%$ for X-VLA and $82\%$ for X-VLA + \method{} over these 50 trials.
Only the policy and adapter run at deployment.
The adapted policies improve by $13$ points on navigation and $16$ points on manipulation.

\begin{table}[t]
\caption{
\textbf{Real-world policy adaptation.}
We compare the original policies with their \method{}-adapted counterparts on two physical platforms.
Navigation reports oracle goal-reaching success on a Unitree Go2 over 100 trials; manipulation reports task completion on a Piper arm over 50 trials across five tasks.
Both metrics are reported as percentages ($\uparrow$).
}
\label{tab:real_robot}
\centering
\small
\setlength{\tabcolsep}{5pt}
\begin{tabular}{@{}lcc@{}}
\toprule
& Navigation$\uparrow$ & Manipulation$\uparrow$ \\
& UniVLA & X-VLA \\
\midrule
Base & 35\% & 66\% \\
\ourrow
\textbf{+ \method{}} & \textbf{48\%} & \textbf{82\%} \\
\bottomrule
\end{tabular}
\end{table}

\begin{figure}[t]
    \centering
    \includegraphics[width=\columnwidth]{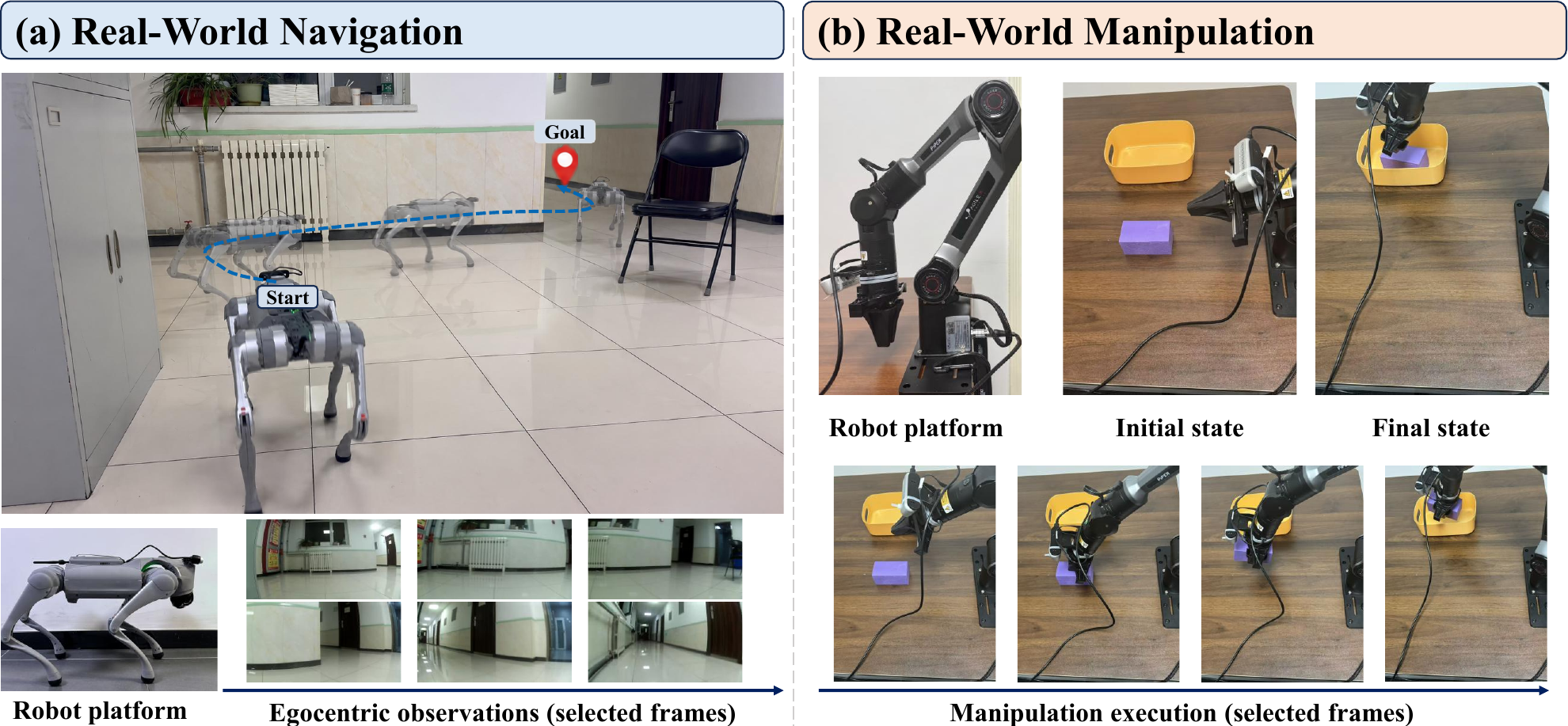}
    \caption{
    \textbf{Real-world navigation and manipulation.}
    (a) A Unitree Go2 performs image-goal navigation using onboard RGB observations.
    The upper image visualizes a representative navigation trajectory, while the lower row shows the robot platform and selected egocentric observations.
    (b) A Piper robotic arm executes a tabletop pick-and-place task.
    The upper row shows the platform and task states before and after execution, and the lower row presents selected frames of the manipulation sequence.
    }
    \label{fig:real_robot}
\end{figure}

\subsection{Ablation and Analysis}
\label{sec:ablation}

We evaluate the effectiveness of the proposed relation-modeling pipeline by comparing it with alternative transition representations, while keeping the latent-action posterior, dynamics model, training objectives, data, and optimization protocol fixed.
The most direct baseline infers $U$ from the same frozen {VGGT-$\Omega$} paired-view representations, without the explicit relation-modeling stage.
We additionally compare against optical flow~\cite{teed2020raft} and differences between independently encoded visual features.
As shown in Table~\ref{tab:relation_ablation}, \method{} improves PG from $0.54$ to $0.58$, $\nas$ from $0.32$ to $0.38$, and RECON ATE from $1.30$ to $0.90$ over the matched paired-view baseline.
In contrast, independent feature differencing yields $0.53$ PG, $0.30$ $\nas$, and $1.50$ ATE, supporting the benefit of the proposed relation-modeling pipeline over direct feature differencing.
The optical-flow baseline also falls short of \method{} in predictive specificity and planning.
These results indicate that the improvement cannot be attributed solely to access to the paired-view prior or visual change features.

Further analyses in Appendix~\ref{app:relation_analysis} examine the learned relation representations.
Linear probes in Table~\ref{tab:relation_content} show improved predictability of physical transition effects ($R^2=0.20$) and reduced source-trajectory predictability ($1.7\%$ vs.\ $17.7\%$) relative to raw paired features.
Appendix~\ref{app:latent_ablate} further analyzes the training objectives: removing action ranking reduces $\nas$ from $0.38$ to $0.25$, while removing KL worsens planning ATE from $0.90$ to $1.06$ despite improving one-step prediction.
These results provide complementary evidence for the roles of relation modeling, action specificity, and latent regularization.

\begin{table}[t]
\caption{
\textbf{Effect of explicit transition-relation modeling.}
We compare \method{} against alternative transition representations while keeping the latent-action posterior, dynamics model, training objectives, data, and optimization protocol fixed.
The paired-view baseline uses the same frozen visual prior but infers $U$ directly without an explicit relation stage.
PG and NAS measure predictive gain and action specificity; ATE and RPE measure navigation planning error.
}
\label{tab:relation_ablation}
\centering
\small
\setlength{\tabcolsep}{4pt}
\begin{tabular}{@{}lcccc@{}}
\toprule
Variant
& \cellcolor{black!8}PG$\uparrow$
& \cellcolor{black!8}$\nas\uparrow$
& \cellcolor{red!10}ATE$\downarrow$
& \cellcolor{red!10}RPE$\downarrow$ \\
\midrule
{VGGT-$\Omega$} pair, no relation-first
& 0.54 & 0.32 & 1.30 & 0.32 \\
Optical flow
& 0.56 & 0.35 & 1.32 & 0.32 \\
Independent feature difference
& 0.53 & 0.30 & 1.50 & 0.35 \\
\ourrow
\textbf{\method{}}
& \textbf{0.58} & \textbf{0.38}
& \textbf{0.90} & \textbf{0.31} \\
\bottomrule
\end{tabular}
\end{table}

\begin{table}[t]
\caption{
\textbf{Transition content versus scene content.}
Motion $R^{2}$ is a ridge readout of $(\Delta x,\Delta y,\Delta\theta)$.
Scene accuracy classifies the source trajectory.
}
\label{tab:relation_content}
\centering
\small
\setlength{\tabcolsep}{6pt}
\begin{tabular}{@{}lcc@{}}
\toprule
Representation & \cellcolor{black!8}Motion $R^{2}\uparrow$ & \cellcolor{red!10}Scene acc.$\downarrow$ \\
\midrule
Paired $\Omega$ features & $-0.04$ & 17.7 \\
Independent $\Delta Z$ & $-0.50$ & 1.9 \\
\ourrow
\textbf{\method{} $R$} & \textbf{0.20} & \textbf{1.7} \\
\bottomrule
\end{tabular}
\end{table}

\section{Conclusion}

Latent-action pretraining typically infers actions directly from paired visual states, implicitly entangling transition representation and action inference.
We argue that action should instead be inferred only after explicitly characterizing what changed between observations.
Based on this principle, \method{} formulates latent-action pretraining as a relation-first problem, yielding a compact action representation that can support predictive dynamics and downstream control.
Experiments across navigation and manipulation, including both in-domain and cross-domain settings, demonstrate the effectiveness and transferability of the learned representation.
Overall, our results suggest that modeling the relation before the action provides a simple and general foundation for latent-action pretraining.

\section*{LLM Usage Statement}

The authors are responsible for the research idea, model design, and all reported results.
Grok 4.7 assisted with code implementation and experiment execution; the authors specified the procedures and verified the outputs.
GPT-6 revised author-written prose, suggested a smaller part of the related work, and gave textual comments on the teaser and method figures, which the authors designed and drew.
Its image generation was used only for a few simple icons.
Real-scene figures, including the real-robot examples, were not produced by a generative model.

{\small
\bibliographystyle{ieeenat_fullname}
\bibliography{main_new}

@String(CVPR = {IEEE Conf. Comput. Vis. Pattern Recog.})

@String(ICCV = {Int. Conf. Comput. Vis.})

@inproceedings{murlabadia2026vjepa21,
  title={{V-JEPA} 2.1: Unlocking dense features in video self-supervised learning},
  author={Mur-Labadia, Lorenzo and Muckley, Matthew and Bar, Amir and Assran, Mido and Sinha, Koustuv and Rabbat, Mike and LeCun, Yann and Ballas, Nicolas},
  booktitle={European Conference on Computer Vision},
  pages={671--689},
  year={2026},
  organization={Springer}
}

@article{assran2025vjepa2,
  title={{V-JEPA} 2: Self-supervised video models enable understanding, prediction and planning},
  author={Assran, Mido and Bardes, Adrien and Fan, David and Garrido, Quentin and Howes, Russell and Muckley, Matthew and Rizvi, Ammar and Roberts, Claire and Sinha, Koustuv and Zholus, Artem and others},
  journal={arXiv preprint arXiv:2506.09985},
  year={2025}
}

@article{wang2026vggtomega,
  title={{VGGT-$\Omega$}},
  author={Wang, Jianyuan and Chen, Minghao and Zhang, Shangzhan and Karaev, Nikita and Sch{\"o}nberger, Johannes and Labatut, Patrick and Bojanowski, Piotr and Novotny, David and Vedaldi, Andrea and Rupprecht, Christian},
  journal={arXiv preprint arXiv:2605.15195},
  year={2026}
}

@inproceedings{edwards2019imitating,
  title={Imitating latent policies from observation},
  author={Edwards, Ashley and Sahni, Himanshu and Schroecker, Yannick and Isbell, Charles},
  booktitle={International conference on machine learning},
  pages={1755--1763},
  year={2019},
  organization={PMLR}
}

@inproceedings{bruce2024genie,
  title={Genie: Generative interactive environments},
  author={Bruce, Jake and Dennis, Michael D and Edwards, Ashley and Parker-Holder, Jack and Shi, Yuge and Hughes, Edward and Lai, Matthew and Mavalankar, Aditi and Steigerwald, Richie and Apps, Chris and others},
  booktitle={Forty-first international conference on machine learning},
  year={2024}
}

@inproceedings{ye2025lapa,
  title={Latent action pretraining from videos},
  author={Ye, Seonghyeon and Jang, Joel and Jeon, Byeongguk and Joo, Se June and Yang, Jianwei and Peng, Baolin and Mandlekar, Ajay and Tan, Reuben and Chao, Yu-Wei and Lin, Bill Yuchen and others},
  booktitle={International Conference on Learning Representations},
  volume={2025},
  year={2025}
}

@article{bu2025univla,
  title={{UniVLA}: Learning to act anywhere with task-centric latent actions},
  author={Bu, Qingwen and Yang, Yanting and Cai, Jisong and Gao, Shenyuan and Ren, Guanghui and Yao, Maoqing and Luo, Ping and Li, Hongyang},
  journal={arXiv preprint arXiv:2505.06111},
  year={2025}
}

@inproceedings{chen2026villax,
  title={{villa-X}: Enhancing latent action modeling in vision-language-action models},
  author={Chen, Xiaoyu and Wei, Hangxing and Zhang, Pushi and Zhang, Chuheng and Wang, Kaixin and Guo, Yanjiang and Yang, Rushuai and Wang, Yucen and Xiao, Xinquan and Zhao, Li and others},
  booktitle={International Conference on Learning Representations},
  volume={2026},
  year={2026}
}

@article{tharwat2025lawm,
  title={Latent action pretraining through world modeling},
  author={Tharwat, Bahey and Nasser, Yara and Abouzeid, Ali and Reid, Ian},
  journal={arXiv preprint arXiv:2509.18428},
  year={2025}
}

@inproceedings{bar2025nwm,
  title={Navigation world models},
  author={Bar, Amir and Zhou, Gaoyue and Tran, Danny and Darrell, Trevor and LeCun, Yann},
  booktitle={2025 IEEE/CVF Conference on Computer Vision and Pattern Recognition (CVPR)},
  pages={15791--15801},
  year={2025},
  organization={IEEE}
}

@inproceedings{shah2023gnm,
  title={{GNM}: A general navigation model to drive any robot},
  author={Shah, Dhruv and Sridhar, Ajay and Bhorkar, Arjun and Hirose, Noriaki and Levine, Sergey},
  booktitle={2023 IEEE International Conference on Robotics and Automation (ICRA)},
  pages={7226--7233},
  year={2023},
  organization={IEEE}
}

@inproceedings{sridhar2024nomad,
  title={{NoMaD}: Goal masked diffusion policies for navigation and exploration},
  author={Sridhar, Ajay and Shah, Dhruv and Glossop, Catherine and Levine, Sergey},
  booktitle={2024 IEEE International Conference on Robotics and Automation (ICRA)},
  pages={63--70},
  year={2024},
  organization={IEEE}
}

@inproceedings{stapf2026come,
  title={Composition of memory experts for diffusion world models},
  author={Stapf, Sebastian and Acuaviva, Pablo and Davtyan, Aram and Favaro, Paolo},
  booktitle={International Conference on Learning Representations},
  volume={2026},
  year={2026}
}

@article{shah2021recon,
  title={Rapid exploration for open-world navigation with latent goal models},
  author={Shah, Dhruv and Eysenbach, Benjamin and Kahn, Gregory and Rhinehart, Nicholas and Levine, Sergey},
  journal={arXiv preprint arXiv:2104.05859},
  year={2021}
}

@article{hirose2023huron,
  title={{SACSoN}: Scalable autonomous control for social navigation},
  author={Hirose, Noriaki and Shah, Dhruv and Sridhar, Ajay and Levine, Sergey},
  journal={IEEE Robotics and Automation Letters},
  volume={9},
  number={1},
  pages={49--56},
  year={2024},
  publisher={IEEE}
}

@inproceedings{triest2022tartandrive,
  title={{TartanDrive}: A large-scale dataset for learning off-road dynamics models},
  author={Triest, Samuel and Sivaprakasam, Matthew and Wang, Sean J and Wang, Wenshan and Johnson, Aaron M and Scherer, Sebastian},
  booktitle={2022 International Conference on Robotics and Automation (ICRA)},
  pages={2546--2552},
  year={2022},
  organization={IEEE}
}

@article{karnan2022scand,
  title={Socially compliant navigation dataset ({SCAND}): A large-scale dataset of demonstrations for social navigation},
  author={Karnan, Haresh and Nair, Anirudh and Xiao, Xuesu and Warnell, Garrett and Pirk, S{\"o}ren and Toshev, Alexander and Hart, Justin and Biswas, Joydeep and Stone, Peter},
  journal={IEEE Robotics and Automation Letters},
  volume={7},
  number={4},
  pages={11807--11814},
  year={2022},
  publisher={IEEE}
}

@inproceedings{hirose2018gonet,
  title={{GONet}: A semi-supervised deep learning approach for traversability estimation},
  author={Hirose, Noriaki and Sadeghian, Amir and V{\'a}zquez, Marynel and Goebel, Patrick and Savarese, Silvio},
  booktitle={2018 IEEE/RSJ international conference on intelligent robots and systems (IROS)},
  pages={3044--3051},
  year={2018},
  organization={IEEE}
}

@inproceedings{garrido2026wildlawm,
  title={Learning latent action world models in the wild},
  author={Garrido, Quentin and Nagarajan, Tushar and Terver, Basile and Ballas, Nicolas and LeCun, Yann and Rabbat, Michael},
  booktitle={Forty-third International Conference on Machine Learning},
  year={2026}
}

@inproceedings{bi2026motus,
  title={Motus: A unified latent action world model},
  author={Bi, Hongzhe and Tan, Hengkai and Xie, Shenghao and Wang, Zeyuan and Huang, Shuhe and Liu, Haitian and Zhao, Ruowen and Feng, Yao and Xiang, Chendong and Rong, Yinze and others},
  booktitle={Proceedings of the IEEE/CVF Conference on Computer Vision and Pattern Recognition},
  pages={35101--35113},
  year={2026}
}

@article{liu2026scar,
  title={{SCAR}: Self-supervised continuous action representation learning},
  author={Liu, Hongjia and Feng, Fan and Fu, Minghao and Wang, Xinyue and Lu, Haofei and Huang, Biwei},
  journal={arXiv preprint arXiv:2605.16412},
  year={2026}
}

@article{huang2026lacwm,
  title={Cross-Embodiment Robot Foundation World Models with Latent Actions},
  author={Huang, Huang and Yenamandra, Sriram and Majumdar, Arjun and Aljalbout, Elie and Nagarajan, Tushar and Yang, Tsung-Yen and Rai, Akshara and Rabbat, Michael and Fei-Fei, Li and Wu, Jiajun and others},
  journal={arXiv preprint arXiv:2610.10846},
  year={2026}
}

@inproceedings{savva2019habitat,
  title={Habitat: A platform for embodied {AI} research},
  author={Savva, Manolis and Kadian, Abhishek and Maksymets, Oleksandr and Zhao, Yili and Wijmans, Erik and Jain, Bhavana and Straub, Julian and Liu, Jia and Koltun, Vladlen and Malik, Jitendra and others},
  booktitle={2019 IEEE/CVF International Conference on Computer Vision (ICCV)},
  pages={9338--9346},
  year={2019},
  organization={IEEE}
}

@article{ramakrishnan2021hm3d,
  title={Habitat-Matterport {3D} dataset ({HM3D}): 1000 large-scale {3D} environments for embodied {AI}},
  author={Ramakrishnan, Santhosh K and Gokaslan, Aaron and Wijmans, Erik and Maksymets, Oleksandr and Clegg, Alex and Turner, John and Undersander, Eric and Galuba, Wojciech and Westbury, Andrew and Chang, Angel X and others},
  journal={arXiv preprint arXiv:2109.08238},
  year={2021}
}

@inproceedings{anderson2018vision,
  title={Vision-and-language navigation: Interpreting visually-grounded navigation instructions in real environments},
  author={Anderson, Peter and Wu, Qi and Teney, Damien and Bruce, Jake and Johnson, Mark and S{\"u}nderhauf, Niko and Reid, Ian and Gould, Stephen and Van Den Hengel, Anton},
  booktitle={2018 IEEE/CVF conference on computer vision and pattern recognition},
  pages={3674--3683},
  year={2018},
  organization={IEEE}
}

@inproceedings{krantz2020beyond,
  title={Beyond the {Nav-Graph}: Vision-and-language navigation in continuous environments},
  author={Krantz, Jacob and Wijmans, Erik and Majumdar, Arjun and Batra, Dhruv and Lee, Stefan},
  booktitle={European Conference on Computer Vision},
  pages={104--120},
  year={2020},
  organization={Springer}
}

@article{lin2025vlnverse,
  title={{VLNVerse}: A benchmark for vision-language navigation with versatile, embodied, realistic simulation and evaluation},
  author={Lin, Sihao and Li, Zerui and Zhao, Xunyi and Zhou, Gengze and Wang, Liuyi and Wei, Rong and Tang, Rui and Li, Juncheng and Wang, Hanqing and Pang, Jiangmiao and others},
  journal={arXiv preprint arXiv:2512.19021},
  year={2025}
}

@article{mees2022calvin,
  title={{CALVIN}: A benchmark for language-conditioned policy learning for long-horizon robot manipulation tasks},
  author={Mees, Oier and Hermann, Lukas and Rosete-Beas, Erick and Burgard, Wolfram},
  journal={IEEE Robotics and Automation Letters},
  volume={7},
  number={3},
  pages={7327--7334},
  year={2022},
  publisher={IEEE}
}

@inproceedings{khanna2024hssd,
  title={Habitat synthetic scenes dataset ({HSSD-200}): An analysis of {3D} scene scale and realism tradeoffs for {ObjectGoal} navigation},
  author={Khanna, Mukul and Mao, Yongsen and Jiang, Hanxiao and Haresh, Sanjay and Shacklett, Brennan and Batra, Dhruv and Clegg, Alexander and Undersander, Eric and Chang, Angel X and Savva, Manolis},
  booktitle={2024 IEEE/CVF Conference on Computer Vision and Pattern Recognition (CVPR)},
  pages={16384--16393},
  year={2024},
  organization={IEEE}
}

@article{khazatsky2024droid,
  title={{DROID}: A large-scale in-the-wild robot manipulation dataset},
  author={Khazatsky, Alexander and Pertsch, Karl and Nair, Suraj and Balakrishna, Ashwin and Dasari, Sudeep and Karamcheti, Siddharth and Nasiriany, Soroush and Srirama, Mohan Kumar and Chen, Lawrence Yunliang and Ellis, Kirsty and others},
  journal={arXiv preprint arXiv:2403.12945},
  year={2024}
}

@article{terver2026physicalplanning,
  title={What drives success in physical planning with joint-embedding predictive world models?},
  author={Terver, Basile and Yang, Tsung-Yen and Ponce, Jean and Bardes, Adrien and LeCun, Yann},
  journal={arXiv preprint arXiv:2512.24497},
  year={2025}
}

@inproceedings{peebles2023dit,
  title={Scalable diffusion models with transformers},
  author={Peebles, William and Xie, Saining},
  booktitle={2023 IEEE/CVF International Conference on Computer Vision (ICCV)},
  pages={4172--4182},
  year={2023},
  organization={IEEE}
}

@article{dosovitskiy2021vit,
  title={An image is worth 16x16 words: Transformers for image recognition at scale},
  author={Dosovitskiy, Alexey and Beyer, Lucas and Kolesnikov, Alexander and Weissenborn, Dirk and Zhai, Xiaohua and Unterthiner, Thomas and Dehghani, Mostafa and Minderer, Matthias and Heigold, Georg and Gelly, Sylvain and Uszkoreit, Jakob and Houlsby, Neil},
  journal={arXiv preprint arXiv:2010.11929},
  year={2020}
}

@inproceedings{teed2020raft,
  title={{RAFT}: Recurrent all-pairs field transforms for optical flow},
  author={Teed, Zachary and Deng, Jia},
  booktitle={European conference on computer vision},
  pages={402--419},
  year={2020},
  organization={Springer}
}

@article{hu2021lora,
  title={{LoRA}: Low-rank adaptation of large language models},
  author={Hu, Edward J and Shen, Yelong and Wallis, Phillip and Allen-Zhu, Zeyuan and Li, Yuanzhi and Wang, Shean and Wang, Lu and Chen, Weizhu},
  journal={arXiv preprint arXiv:2106.09685},
  year={2021}
}

@article{rombach2022ldm,
  title={High-Resolution Image Synthesis with Latent Diffusion Models},
  author={Rombach, Robin and Blattmann, Andreas and Lorenz, Dominik and Esser, Patrick and Ommer, Bj{\"o}rn},
  journal={arXiv preprint arXiv:2112.10752},
  year={2022}
}

@article{zheng2025x,
  title   = {{X-VLA}: Soft-prompted transformer as scalable cross-embodiment vision-language-action model},
  author  = {Zheng, Jinliang and Li, Jianxiong and Wang, Zhihao and Liu, Dongxiu and Kang, Xirui and Feng, Yuchun and Zheng, Yinan and Zou, Jiayin and Chen, Yilun and Zeng, Jia and others},
  journal = {arXiv preprint arXiv:2510.10274},
  year    = {2025}
}

@article{battaglia2018relational,
  title={Relational inductive biases, deep learning, and graph networks},
  author={Peter W. Battaglia and Jessica B. Hamrick and Victor Bapst and Alvaro Sanchez-Gonzalez and Vin{\'i}cius Flores Zambaldi and Mateusz Malinowski and Andrea Tacchetti and David Raposo and Adam Santoro and Ryan Faulkner and Çaglar G{\"u}lçehre and H. Francis Song and Andrew J. Ballard and Justin Gilmer and George Edward Dahl and Ashish Vaswani and Kelsey R. Allen and Charlie Nash and Victoria Langston and Chris Dyer and Nicolas Manfred Otto Heess and Daan Wierstra and Pushmeet Kohli and Matthew M. Botvinick and Oriol Vinyals and Yujia Li and Razvan Pascanu},
  journal={ArXiv},
  year={2018},
  volume={abs/1806.01261},
  url={https://api.semanticscholar.org/CorpusID:46935302}
}

@article{kipf2020contrastive,
  title={Contrastive Learning of Structured World Models},
  author={Thomas Kipf and Elise van der Pol and Max Welling},
  journal={ArXiv},
  year={2019},
  volume={abs/1911.12247},
  url={https://api.semanticscholar.org/CorpusID:208310100}
}

@article{alemi2017deep,
  title={Deep Variational Information Bottleneck},
  author={Alexander A. Alemi and Ian Fischer and Joshua V. Dillon and Kevin P. Murphy},
  journal={ArXiv},
  year={2017},
  volume={abs/1612.00410},
  url={https://api.semanticscholar.org/CorpusID:204922497}
}

@article{wang2026flam,
  title={Factored Latent Action World Models},
  author={Zi-Zhao Wang and Chang Shi and Jiaheng Hu and Kevin Rohling and Roberto Mart{\'i}n-Mart{\'i}n and Amy Zhang and Peter Stone},
  journal={ArXiv},
  year={2026},
  volume={abs/2602.16229},
  url={https://api.semanticscholar.org/CorpusID:285725804}
}

@inproceedings{chen2026cyclemimic,
  title={Learning a Unified Latent Action Space from Videos with Action-centric Cycle Consistency},
  author={Guang-Yan Chen and Qi Shao and Te Cui and Zi-Chen Zhou and Wei-Xin Mao and Luo-Jie Yang and Mei-Ling Wang and Yi Yang and Hua Chen and Yu-Feng Yue},
  booktitle={Computer Vision and Pattern Recognition},
  year={2026},
  url={https://api.semanticscholar.org/CorpusID:295645447}
}
}

\clearpage
\appendix
\section{Theoretical Foundation of Relation-First Predictive Abstraction}
\label{app:predictive_theory}

We provide a formal connection between the
relation-first factorization and its practical
predictive learning objective.
The analysis distinguishes information loss
during relation construction from information
loss during latent action abstraction.

\subsection{A Predictive Abstraction Principle}

The latent action must retain sufficient information about the observed transition to predict its consequence, despite having a substantially more compact representation than the relation $R$.
We formalize this requirement through predictive information preservation.
Let $H=\mathcal H_t$ and $Z^{+}=Z_{t+\tau}$, with $V$ denoting the paired-view evidence introduced in Sec.~\ref{sec:formulation}.

Consider an ideal probabilistic predictor evaluated under conditional log-loss.
Its minimum achievable prediction risk from $(H,U)$ is $H(Z^{+}\mid H,U)$.
The richer evidence $(H,V)$ provides a reference predictor with optimal risk $H(Z^{+}\mid H,V)$.
Their difference,
$H(Z^{+}\mid H,U)-H(Z^{+}\mid H,V)$,
therefore measures the predictive information lost when the paired-view evidence is replaced by the latent action.
Minimizing this gap encourages $U$ to retain the information needed to explain the future.

Under the information flow $V\rightarrow R\rightarrow U$ conditioned on $H$, this predictive gap decomposes as
\begin{equation}
\begin{aligned}
&H(Z^{+}\mid H,U)-H(Z^{+}\mid H,V)\\
={}&
\underbrace{I(Z^{+};V\mid H,R)}_{\text{relation insufficiency}}
+
\underbrace{I(Z^{+};R\mid H,U)}_{\text{action insufficiency}}.
\end{aligned}
\label{eq:predictive_decomposition}
\end{equation}

The decomposition reveals two potential sources of information loss.
The first term measures predictive evidence discarded when $R$ is constructed from $V$; the second measures additional predictive information lost when $R$ is compressed into $U$.
Both must be small for the final action representation to preserve the predictive content available along the relation-first pathway.

However, predictive sufficiency alone does not favor compact actions: an unconstrained $U$ could retain unnecessary relation-specific information.
We therefore combine predictive preservation with an information-rate penalty, yielding the ideal abstraction objective
\begin{equation}
\boxed{
\begin{aligned}
\mathcal J_{\mathrm{RELACT}}
={}&
I(Z^{+};V\mid H,R)\\
&+
I(Z^{+};R\mid H,U)\\
&+
\beta_{\mathrm{IB}}I(U;R\mid H).
\end{aligned}
}
\label{eq:relation_action_principle}
\end{equation}

The first two terms encourage predictive sufficiency across relation construction and action abstraction; the third limits how much information the latent action retains from the relation.
Importantly, this objective characterizes the desired representation rather than specifying three separately optimized losses.
The relation-first factorization provides the architectural inductive bias, while prediction and information regularization jointly supervise the entire pathway.
The next subsection derives this objective from conditional log-loss and a posterior-to-prior KL penalty.

\subsection{Two-Stage Predictive Information Preservation}

Let $V$ be the paired-view visual evidence,
$R=\mathcal R_\eta(V)$ the learned transition relation,
$H$ the available visual history, and
$Z^{+}$ the future visual representation.
The latent action is sampled according to
$q_\phi(U\mid Z_t,R)$, where $Z_t$ is
determined by $H$.

The resulting information flow satisfies
\begin{equation}
Z^{+}\longleftrightarrow(H,V)
\longrightarrow R
\longrightarrow U,
\label{eq:app_information_flow}
\end{equation}
with the final transitions interpreted
conditionally on $H$.

We assume that the required conditional
information quantities and log-loss risks
are finite.

\begin{proposition}[Relation--action predictive decomposition]
\label{prop:relation_action_decomp}
The predictive information loss from $V$ to $U$
decomposes as
\begin{equation}
\boxed{
\begin{aligned}
&I(Z^{+};V\mid H)-I(Z^{+};U\mid H)\\
&=
I(Z^{+};V\mid H,R)
+
I(Z^{+};R\mid H,U).
\end{aligned}
}
\label{eq:app_two_stage_decomp}
\end{equation}

Moreover, let $p_\theta(Z^{+}\mid H,U)$ denote
a probabilistic dynamics model and
$p_0(U)$ a fixed action prior.
Define
\begin{equation}
\begin{aligned}
\mathcal L_{\mathrm{var}}
={}&
\mathbb E[-\log p_\theta(Z^{+}\mid H,U)]\\
&+
\beta_{\mathrm{IB}}\,
\mathbb E_{H,R}
D_{\mathrm{KL}}
(q_\phi(U\mid H,R)\Vert p_0(U)).
\end{aligned}
\end{equation}

Then
\begin{equation}
\boxed{
\begin{aligned}
\mathcal L_{\mathrm{var}}
={}&
H(Z^{+}\mid H,V)
+\mathcal J_{\mathrm{RELACT}}\\
&+\mathcal E_{\mathrm{model}}
+\beta_{\mathrm{IB}}\mathcal E_{\mathrm{prior}},
\end{aligned}
}
\label{eq:app_variational_relation}
\end{equation}
where
\begin{align}
\mathcal E_{\mathrm{model}}
&=
\mathbb E_{H,U}
D_{\mathrm{KL}}
\big(
p(Z^{+}\mid H,U)
\Vert p_\theta(Z^{+}\mid H,U)
\big),\\
\mathcal E_{\mathrm{prior}}
&=
\mathbb E_H
D_{\mathrm{KL}}
\big(q_\phi(U\mid H)\Vert p_0(U)\big),
\end{align}
and both terms are nonnegative.
\end{proposition}

\begin{proof}
Since $R$ is a function of $V$,
\begin{equation}
H(Z^{+}\mid H,R)-H(Z^{+}\mid H,V)
=
I(Z^{+};V\mid H,R).
\end{equation}

Because the sampling of $U$ depends on
$(H,R)$ but not additionally on $Z^{+}$,
we have $Z^{+}\perp U\mid(H,R)$ and hence
\begin{equation}
H(Z^{+}\mid H,U)-H(Z^{+}\mid H,R)
=
I(Z^{+};R\mid H,U).
\end{equation}

Adding the two identities gives
\begin{equation}
\begin{aligned}
&H(Z^{+}\mid H,U)-H(Z^{+}\mid H,V)\\
&=
I(Z^{+};V\mid H,R)
+
I(Z^{+};R\mid H,U).
\end{aligned}
\end{equation}
Expressing the entropy difference
in terms of mutual information proves
Eq.~\eqref{eq:app_two_stage_decomp}.

For the predictive loss,
the conditional cross-entropy identity gives
\begin{equation}
\begin{aligned}
&\mathbb E[-\log p_\theta(Z^{+}\mid H,U)]\\
&=
H(Z^{+}\mid H,U)
+\mathcal E_{\mathrm{model}}.
\end{aligned}
\end{equation}

For the latent-action KL penalty,
introduce the aggregated posterior
$q_\phi(U\mid H)$.
Expanding the KL divergence yields
\begin{equation}
\begin{aligned}
&\mathbb E_{H,R}
D_{\mathrm{KL}}
(q_\phi(U\mid H,R)\Vert p_0(U))\\
&=
I(U;R\mid H)
+\mathcal E_{\mathrm{prior}}.
\end{aligned}
\end{equation}

Substituting these two decompositions
into $\mathcal L_{\mathrm{var}}$
and using the first part of the proposition
establishes
Eq.~\eqref{eq:app_variational_relation}.
\end{proof}

\subsection{Implications for \method{}}

Proposition~\ref{prop:relation_action_decomp}
connects the two stages of \method{}
to a unified predictive objective.

\noindent\textbf{Relation construction.}
The term $I(Z^{+};V\mid H,R)$ characterizes
predictive information in the paired-view
evidence that is not retained by $R$.
This motivates constructing the relation
from jointly contextualized observations
rather than treating independently encoded
state differences as the only available
transition evidence.
The paired-view construction is an inductive
bias for retaining useful relational
information.

\noindent\textbf{Action abstraction.}
The term $I(Z^{+};R\mid H,U)$ characterizes
additional predictive information lost
when compressing the relation into $U$.
The state-conditioned posterior and forward
dynamics jointly learn this abstraction,
while the latent-action information rate
is constrained through KL regularization.

\noindent\textbf{Connection to optimization.}
For a fixed pretrained paired-view
representation, $H(Z^{+}\mid H,V)$ is independent
of the learnable relation projection,
action posterior, and dynamics model.
Equation~\eqref{eq:app_variational_relation}
therefore gives a variational interpretation
of their joint predictive optimization.

In practice, \method{} uses feature-space
cosine distance and MSE rather than exact
conditional log-likelihood.
Under a fixed-variance Gaussian predictive
model, squared error is proportional to
negative log-likelihood up to constants;
the cosine term is an additional surrogate.

The action-intervention ranking loss and
detached rollout loss further encourage
transition-specific action usage and
multi-step predictive consistency.

Relation-first learning organizes transition
evidence explicitly and then abstracts it into
a predictive latent action.
The directed relational graph in
Sec.~\ref{sec:relation} is one concrete
realization of this principle.
Its shared scene queries, directed pairwise
edges, asymmetric bilinear interactions, and
fixed image-center reference node specify how
that evidence is assembled into $R$.
Table~\ref{tab:relation_ablation} compares the
full pipeline with other transition
representations.
How much any single graph operator contributes
is a finer architecture-level question,
separate from this framework-level comparison.

\section{Implementation Details of Latent-Action Reuse}
\label{app:reuse}

We provide additional details on how the pretrained
latent action space is connected to executable native
controls and existing embodied policies.
In both settings, the pretrained relation encoder
and action posterior remain frozen.

\subsection{Native-Action Grounding and Planning}
\label{app:grounding}

\noindent\textbf{Grounding native controls.}\ 
For each embodiment $e$, we construct a dataset of
labeled transitions
$\mathcal D_e=\{(I_t,a_t^e,I_{t+\tau})\}$,
where $a_t^e$ denotes the native control associated
with the corresponding transition interval.

The frozen relation-to-action posterior serves as
a transition-conditioned teacher:
\begin{equation}
q_\phi(U\mid Z_t,R_{t,t+\tau})
=
\mathcal N(\mu_q,\operatorname{diag}\sigma_q^2).
\end{equation}

We train a lightweight state-conditioned grounding
model with a diagonal Gaussian output:
\begin{equation}
p_e(U\mid Z_t,a_t^e)
=
\mathcal N\left(
\mu_e(Z_t,a_t^e),
\operatorname{diag}\sigma_e^2(Z_t,a_t^e)
\right).
\label{eq:app_ground_dist}
\end{equation}

The grounding objective is
\begin{equation}
\mathcal L_{\mathrm{ground}}
=
\mathbb E_{\mathcal D_e}
\left[
D_{\mathrm{KL}}\left(
\operatorname{sg}[q_\phi(U\mid Z_t,R_{t,t+\tau})]
\Vert
p_e(U\mid Z_t,a_t^e)
\right)
\right],
\label{eq:app_ground_loss}
\end{equation}
where $\operatorname{sg}$ indicates that the teacher
distribution is not updated.
The reported grounding network applies two pre-norm
self-attention blocks with eight heads to the
$768$-dimensional visual tokens, embeds the native
control with a three-layer MLP of width $768$, and
reads out the $128$-dimensional latent mean by
cross-attention and a linear head.
It is fit by squared error to the frozen teacher
mean $\mu_q$, using AdamW with learning rate
$10^{-3}$, weight decay $0.04$, a global batch of
$256$, and $3{,}000$ steps.

The teacher observes the completed transition
through $R_{t,t+\tau}$, whereas the grounding model
receives only the current state and the corresponding
native control.
It therefore learns to predict the latent transition
effect associated with a candidate control without
requiring future observations at inference.

We condition the grounding model on $Z_t$ to obtain
a local state-dependent control-to-latent interface.
The dynamics model separately uses the visual history
$\mathcal H_t$ to account for temporal context during
future-state prediction.
This separation maintains a lightweight grounding
interface while preserving history-conditioned dynamics.

\noindent\textbf{Model-based planning.}\ 
At inference, planning is performed over candidate
sequences in the native-action space.

For each candidate sequence, the grounding model
maps its native controls into latent actions.
The frozen dynamics model then predicts the
corresponding future representations, using the
predicted states and updated visual history
throughout the planning horizon.

The resulting trajectories are evaluated using
the task-specific goal cost.
The planner ranks candidate native-action
sequences according to their predicted consequences,
rather than directly decoding latent actions into
native controls.

For RECON evaluation, we follow the top-$K_{\mathrm{top}}$
mean-action execution protocol with $K_{\mathrm{top}}=5$:
the last-step native controls of the five highest-ranked
candidate sequences are averaged to obtain the
executed control, which is then repeated across the
planning horizon.

Thus, native-action grounding supports the complete
planning pathway from executable control proposals,
through latent prediction and candidate evaluation,
back to executable native controls.

\subsection{Transition-Aware Policy Adaptation}
\label{app:policy_adapter}

\noindent\textbf{Adapter and transition supervision.}\ 
Consider a pretrained policy $\pi$ that produces
an internal representation $h_t$ and original
action logits $\ell_t$.
We freeze the base policy and introduce a
lightweight adapter $A_\psi$ to predict
$\widehat U_t=A_\psi(h_t)$.

During training, an expert transition provides
the target latent action
\begin{equation}
U_t^\star
=
\mu_\phi\left(
Z_t,R_{t,t+\tau}
\right),
\label{eq:app_policy_teacher}
\end{equation}
where the relation is constructed from the
current and expert future observations over
the corresponding prediction horizon.

The adapter is trained to infer this latent
transition from the policy representation,
without accessing the expert future observation.

The predicted action is mapped into a bounded
residual correction of the original policy
output, as defined in
Eq.~\eqref{eq:policy_residual}.
The final layer of the residual branch is
zero-initialized, ensuring that the adapted
policy initially reproduces the frozen base policy.

\noindent\textbf{Training objective.}\ 
We jointly optimize the downstream task objective,
latent-action alignment, and residual regularization:
\begin{equation}
\begin{aligned}
\mathcal L_{\mathrm{adapt}}
={}&
\mathcal L_{\mathrm{task}}
(\ell_t',a_t^\star)\\
&+
\lambda_U
\|\widehat U_t-U_t^\star\|_2^2
+
\lambda_{\mathrm{reg}}\mathcal L_{\mathrm{reg}}.
\end{aligned}
\label{eq:app_policy_loss}
\end{equation}

Here, $\mathcal L_{\mathrm{task}}$ preserves the
original downstream action-learning objective.
The latent-alignment term supervises the adapter
using the pretrained action representation of
the expert transition, while
$\mathcal L_{\mathrm{reg}}$ constrains deviations
from the frozen base policy.
Each adapter uses two MLPs of hidden width $256$
with one GELU: the first maps the frozen policy
feature ($1024$ dimensions on R2R, $512$ on CALVIN)
to a $128$-dimensional latent action, and the second
maps that latent action to the residual, with a
zero-initialized last layer.
We set $\lambda_U=0.1$ on a training-set-standardized
latent target, and define $\mathcal L_{\mathrm{reg}}$
as the mean squared residual, with
$\lambda_{\mathrm{reg}}=0.01$ on R2R and $0.1$ on
CALVIN.
R2R also adds a KL of weight $0.5$ between the
adapted and frozen action distributions and clips
the combined logit change at $0.25$; CALVIN bounds
the residual by $0.1\tanh(\cdot)$.
R2R trains the adapter with AdamW at learning rate
$10^{-4}$, weight decay $10^{-4}$, and the final
action projection at $10^{-5}$, using batch size
$128$ for $8{,}000$ steps.
CALVIN trains only the adapter with AdamW at
$3\times10^{-4}$, the same weight decay, batch size
$64$, and $2{,}000$ steps.

The alignment supervision is applied over the
same temporal horizon used to construct the
teacher latent action, ensuring consistency
between the predicted transition and its target.

\noindent\textbf{Inference.}\ 
At deployment, the frozen relation-to-action
teacher and expert future observations are removed.
The adapter predicts $\widehat U_t$ directly
from $h_t$ and applies the residual correction
before the policy's original action-selection
procedure.

Unlike model-based planning, policy adaptation
does not require online latent dynamics rollouts.
The pretrained action representation is reused
as transition-aware supervision within the
existing policy architecture.

\section{Evaluation Protocol}
\label{app:eval}

\paragraph{Metrics.}
Navigation planning reports ATE and RPE. Lower values are better for both metrics.
ATE is the root-mean-square Euclidean distance between predicted and reference waypoint positions, without rigid alignment or scale correction.
RPE is the root-mean-square translational error between consecutive waypoints at a one-step interval.
DROID reports the translational error
\begin{equation}
    \Delta xyz =
    \left\|
    \sum_{i=t}^{t+H-1} a_i^{\mathrm{plan}}
    -
    \sum_{i=t}^{t+H-1} a_i^{\mathrm{gt}}
    \right\|_1 ,
\end{equation}
where $a_i^{\mathrm{plan}}$ and $a_i^{\mathrm{gt}}$ are the planned and ground-truth translational actions and $H=3$.
The reported number is the average over 64 episodes. Lower is better.

On R2R-CE, oracle success is the fraction of episodes in which the agent comes within $3$\,m of the goal at any point.
On CALVIN, 5-task chain success is the fraction of evaluation sequences that complete five tasks in a row.
Average length is the mean number of consecutively completed tasks and has a maximum of 5.
Higher values indicate better performance for all three metrics.
Real-robot evaluation reports task success over repeated trials.

\paragraph{Splits and action interfaces.}
RECON, TartanDrive, SCAND, and Go Stanford use the official 100-trajectory evaluation splits.
The high-resolution HuRoN release used by prior tables is not public, so HuRoN, including the GNM, NoMaD, and NWM numbers on that column, uses our own split.
TartanDrive GNM, NoMaD, and NWM are recomputed on the same official 100 trajectories.
Go Stanford is excluded from world-model training.
Only a scalar waypoint-scale calibration is fit on its training split.
Before the shared navigation normalization, metric waypoint spacing is 0.25\,m on RECON, 0.255\,m on HuRoN, 0.38\,m on SCAND, and 0.72\,m on TartanDrive.
Those waypoints are then normalized with one shared minimum and maximum.
The in-domain model uses one checkpoint for all five navigation domains.
On DROID, the interface predicts 4\,fps translational displacements in meters, and the reported error is the $\Delta xyz$ above.

\paragraph{Image-goal navigation.}
Following NWM~\cite{bar2025nwm} and CoME~\cite{stapf2026come}, out-of-domain transfer evaluates on 100 sampled RECON trajectories.
Given a current observation and a goal image, the model plans an eight-step trajectory at four frames per second with 120 candidates and the mean of the top-5 last-step actions, matching the NWM search budget.
In-domain navigation scores those candidates by the predicted waypoint trajectory together with the latent-action rollout, not by pixel LPIPS.
Out-of-domain transfer uses the same search with a world model pretrained on HM3D, VLNVerse, and HSSD.
The world-model weights are not updated on RECON, while a RECON controller and geometry readout are trained on the RECON training split.

\paragraph{DROID planning.}
We follow the protocol of Wild-LAWM~\cite{garrido2026wildlawm}.
Each evaluation instance consists of a current observation and a goal observation separated by a planning horizon of $H=3$ steps.
Evaluation uses 64 episodes sampled from 16 held-out DROID validation videos, with clips of $H+1=4$ frames sampled at $4$\,fps.
Given an initial representation $Z_t$ and goal representation $Z_g$,
we search for an action sequence $a_{t:t+H-1}$.
The planning cost is
\begin{equation}
    C(Z_t,a_{t:t+H-1},Z_g)
    =
    \|Z_g-\widehat Z_{t+H}\|_2 ,
\end{equation}
where the predicted states are obtained by recursively rolling out the latent dynamics using the action-to-$U$ interface.
CEM uses 300 candidate sequences, 10 elites, and 15 optimization iterations.
The reported error is the $\Delta xyz$ defined above.

\paragraph{Policy adaptation.}
The policy experiments freeze a pretrained UniVLA policy and train only the lightweight adapter and residual head in Sec.~\ref{sec:reuse}.
The residual baseline uses that same capacity and the downstream action loss, without latent-transition supervision.
\method{} adds supervision from the frozen posterior on expert transitions.
At deployment the teacher path is removed, so the adapter reads only the current policy representation.
R2R-CE reports the oracle success defined above, and CALVIN reports the chain success and average length defined above.

\section{Pretraining Data, Optimization, and Cost}
\label{app:pretrain}

Each pretraining regime learns its own posterior and dynamics model.
The in-domain navigation model, the out-of-domain indoor model, and the DROID manipulation model are not initialized from one another, and their latent spaces are not shared.
All three freeze the V-JEPA~2.1 encoder and VGGT-$\Omega$.
RGB clips contain 8 frames.
The encoder input is $384\times384$, and the paired-view prior uses $512\times512$ inputs.
The relation follows Eq.~\eqref{eq:relation_slots}--\eqref{eq:relation}. Source-anchored scene queries extract corresponding slots from the frozen paired-view tokens, and a learned encoder aggregates changes in their directed visual and image-center-relative configurations, including a fixed image-plane reference node. That node is a fixed image-center anchor, not a recovered camera pose, robot state, or physical motion scale. The slot extractor and relation encoder are trainable; the paired-view encoder remains frozen.
The latent action is a 128-dimensional Gaussian conditioned on $(Z_t,R)$.
The dynamics model conditions on a visual history of length $h=4$ and is also trained with detached closed-loop rollouts of $L_{\mathrm{roll}}=4$ steps.

Optimization is shared across these runs.
The prediction, ranking, KL, and rollout weights are
$\lambda_{\mathrm{mse}}=1$, $\gamma=1$, $\beta=10^{-4}$, and $\lambda_{\mathrm{roll}}=1$, with ranking margin $m=0.1$.
The KL weight warms up for 200 steps and ramps for 400 steps.
We use AdamW with learning rate $3\times10^{-4}$, weight decay $0.01$, and gradient clipping at $1.0$.
The in-domain navigation run uses a global batch of 8 for 40{,}000 updates.
The out-of-domain indoor run uses the same batch for 32{,}000 updates.

\paragraph{In-domain navigation.}
Training uses RECON, HuRoN, TartanDrive, and SCAND.
Each source is capped at 25{,}000 clips.
Go Stanford is not included in this training mix.

\paragraph{Out-of-domain transfer.}
Table~\ref{tab:ood_pretrain} accounts for the indoor world model that is later transferred to RECON, and for the Wild-LAWM pretraining budget reported in that paper.
The indoor mix contains 49{,}600 HM3D clips, 40{,}000 VLNVerse clips, and 39{,}680 HSSD clips, 129{,}280 clips in total.
At 32{,}000 updates and batch 8, this is $2.56\times10^{5}$ clip-updates.
Wild-LAWM trains on YT-Temporal-1B with 16-frame clips at 4\,fps, for 30{,}000 updates at batch 1{,}024, about $3.07\times10^{7}$ clip-updates.
\method{} uses fewer clip-updates and still obtains the lower RECON error.

\begin{table}[t]
\caption{
\textbf{Out-of-domain world-model pretraining} versus Wild-LAWM.
Wild-LAWM numbers are taken from their paper.
}
\label{tab:ood_pretrain}
\centering
\small
\setlength{\tabcolsep}{3pt}
\begin{tabular}{@{}lcc@{}}
\toprule
& \cellcolor{blue!12}\method{} & \cellcolor{black!6}Wild-LAWM~\cite{garrido2026wildlawm} \\
\midrule
Source & HM3D, VLNVerse, HSSD & YT-Temporal-1B \\
Domain & indoor RGB clips & in-the-wild video \\
Updates & $32{,}000$ & $30{,}000$ \\
Batch & $8$ & $1{,}024$ \\
Clip-updates & $2.56{\times}10^{5}$ & $3.07{\times}10^{7}$ \\
\bottomrule
\end{tabular}
\end{table}

\paragraph{DROID.}
Manipulation planning uses a separate DROID pretraining run with the same objective.
Native actions are Cartesian translations in meters at 4\,fps.
This run is not initialized from either navigation checkpoint.

\paragraph{Pretraining budget.}
Table~\ref{tab:pretrain_cost} places the in-domain navigation run next to the training setups reported for NWM and Wild-LAWM.
Trainable parameters of \method{} are the scene-slot extractor, the relation encoder, $q_\phi$, and $F_\theta$.
The frozen parameters are V-JEPA~2.1 and VGGT-$\Omega$.
Caching $\Omega$ tokens removes the paired-view forward from the step.
With cached $\Omega$, one step uses 6.1\,GiB and 0.18\,s of forward time.
The 40{,}000-update run took 30.5 hours on two GPUs, $2\times 30.5$ GPU-hours.
NWM trains a 1B-parameter CDiT-XL~\cite{peebles2023dit} with a frozen Stable Diffusion VAE~\cite{rombach2022ldm}.
Its effective batch is 4{,}096, from a batch of 1{,}024 with four navigation goals, on eight machines of eight H100 GPUs.
The paper does not report the number of updates.
The authors later state that the known-environment training takes less than one day on these 64 H100 GPUs, so the budget is less than 1{,}536 GPU-hours.
Wild-LAWM trains a ViT-L predictor on frozen V-JEPA~2-L features for 30{,}000 updates at batch 1{,}024, reported as about 12 hours on 64 H100 GPUs, $64\times 12$ GPU-hours.
The hardware, batch, and data differ, so the table records the published setups.
CoME finetunes a DiT-B~\cite{peebles2023dit} and adapts a frozen NWM with LoRA~\cite{hu2021lora}, and does not report a separate pretraining budget.

\begin{table}[t]
\caption{
\textbf{Pretraining budget.}
NWM and Wild-LAWM model and batch entries are taken from their papers.
Training is the number of GPUs times wall-clock hours.
The products are 61, less than 1{,}536, and 768 GPU-hours.
The NWM bound is the authors' later statement of less than one day on 64 H100 GPUs.
The paper still does not report the number of updates.
The reported memory usage (6.1\,GiB) and forward time (0.18\,s per step) are measured with cached $\Omega$ features.
}
\label{tab:pretrain_cost}
\centering
\small
\setlength{\tabcolsep}{3pt}
\begin{tabular}{@{}lccc@{}}
\toprule
& \cellcolor{blue!12}\method{} & \cellcolor{black!6}NWM~\cite{bar2025nwm} & \cellcolor{black!6}Wild-LAWM~\cite{garrido2026wildlawm} \\
\midrule
Trainable & 44.8M & 1B & ViT-L \\
Frozen & 1.23B & SD VAE & V-JEPA~2-L \\
Updates & $40{,}000$ & -- & $30{,}000$ \\
Batch & $8$ & $4{,}096$ & $1{,}024$ \\
Training & $2{\times}30.5$\,h & $64$ GPUs $\times$ $<$24\,h & $64{\times}12$\,h \\
\bottomrule
\end{tabular}
\end{table}

In the transition-representation comparison of Table~\ref{tab:relation_ablation}, the variants have comparable trainable capacity: $44.1$M for the paired-view baseline, $43.3$M for optical flow, $43.5$M for independent feature difference, and $44.8$M for \method{}.

\section{Real-Robot Protocol}
\label{app:real}

Navigation deploys the UniVLA policy from Sec.~\ref{sec:policy_exp}, and manipulation deploys {X-VLA}~\cite{zheng2025x}.
The \method{} posterior supervises the adapter only during training.
At test time the posterior is removed, and neither platform receives future observations.

\paragraph{Navigation.}
Navigation runs on a Unitree Go2 quadruped with an Intel RealSense D435i RGB-D camera.
The camera is rigidly mounted, and its optical center is approximately $0.42$\,m above the ground.
High-level heading and forward commands from the policy are sent to the vendor locomotion controller, and no target-domain fine-tuning is used.
A commanded $15^\circ$ turn or $0.25$\,m forward move is subject to hardware-level execution error.
The evaluation uses eight scenes, five indoor and three outdoor, and a shared pool of 100 image-goal tasks.
UniVLA and UniVLA + \method{} each execute every task once under the same scene conditions.
A trial counts as an oracle success when the executed trajectory passes through the goal location, whether or not the robot stops there.
Rollout videos from the two methods are anonymized and shuffled before scoring, and the scorer does not know which method produced a video.
For each task we record the goal image, the executed action trajectory, egocentric images, and a scene-level trajectory video.
These recordings are withheld in the submitted version because of the double-blind policy and will be released after acceptance.

\begin{figure}[t]
\centering
\caption{
Per-task manipulation success rates on the Piper arm.
For each task, the policy is fine-tuned using 30 demonstrations and evaluated over 10 trials.
The total is over all 50 trials.
}
\label{fig:real_manip}
\begin{tikzpicture}
\begin{axis}[
    xbar,
    bar width=5.2pt,
    scale only axis,
    width=0.56\columnwidth,
    height=4.45cm,
    xmin=0, xmax=118,
    xtick={0,20,40,60,80,100},
    xlabel={Success rate (\%)},
    xlabel style={font=\small},
    ytick={6,5,4,3,2,0.35},
    yticklabels={
      {Cube into box},
      {Stack blocks},
      {Cylinder into box},
      {Cube onto coaster},
      {Select the correct cube},
      {Total},
    },
    yticklabel style={font=\scriptsize},
    xticklabel style={font=\scriptsize},
    ymin=-0.55, ymax=6.75,
    nodes near coords,
    every node near coord/.append style={font=\scriptsize, /pgf/number format/fixed, /pgf/number format/precision=0},
    legend style={font=\scriptsize, at={(0.5,1.04)}, anchor=south, legend columns=-1, draw=none, fill=none, inner sep=1pt},
    legend cell align=left,
    axis x line*=bottom,
    axis y line*=left,
    xmajorgrids=true,
    grid style={black!12},
    tick style={black!55},
    axis line style={black!70},
    clip=false,
    enlarge y limits=false,
]
\addplot[fill=black!18, draw=black!45, area legend] coordinates {
    (80,6) (60,5) (50,4) (80,3) (60,2) (66,0.35)
};
\addplot[fill=tblaccent, draw=tblaccent!75!black, area legend] coordinates {
    (90,6) (80,5) (70,4) (90,3) (80,2) (82,0.35)
};
\legend{X-VLA, $+$ \method{}}
\end{axis}
\end{tikzpicture}
\end{figure}
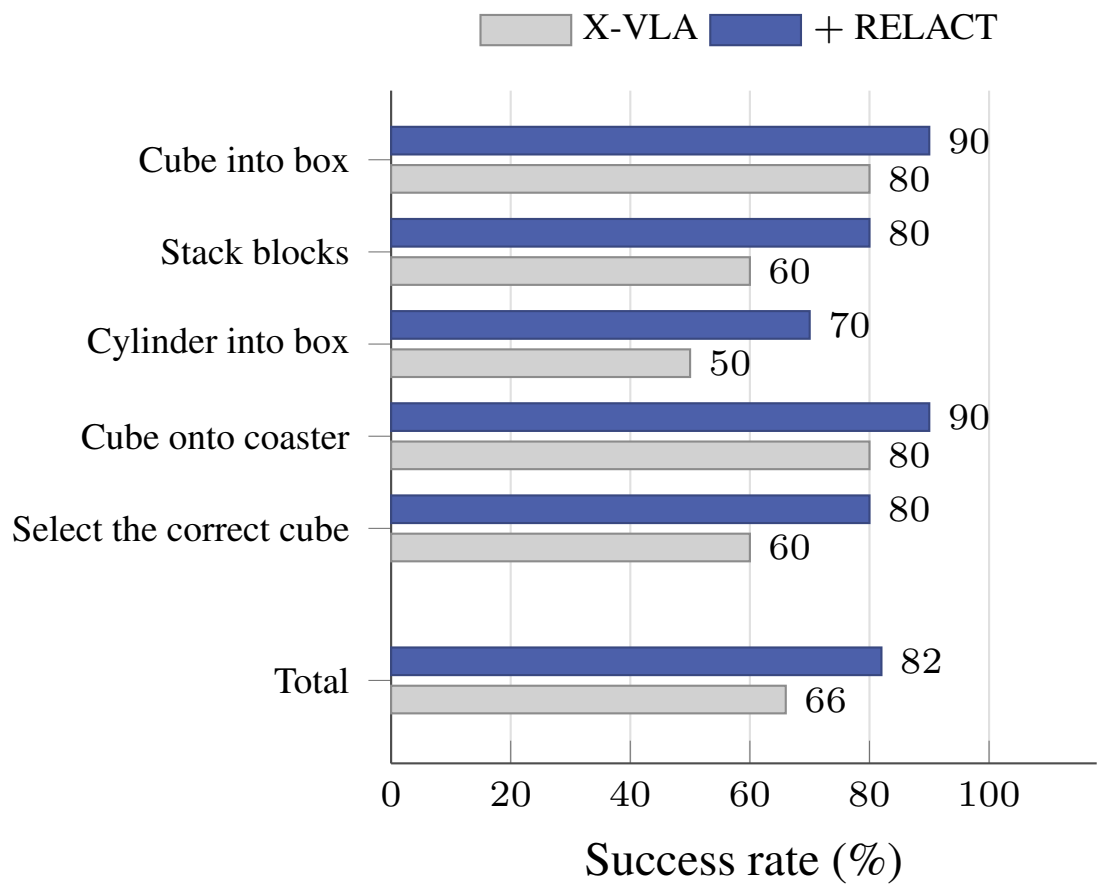

\paragraph{Manipulation.}
Manipulation uses X-VLA on a six-axis Piper arm.
The policy is fine-tuned separately for each of the five tasks using 30 demonstrations.
The tasks are placing a cube into a storage box, stacking blocks, placing a small cylinder into a box, placing a cube on a coaster, and selecting the correct cube among several and placing it on a coaster.
For each experiment we retain the primary-view and third-person image sequences and videos.
Fig.~\ref{fig:real_manip} shows the success rates.

\section{Understanding the Learned Transition Relation}
\label{app:relation_analysis}

\paragraph{Prediction gain and action specificity.}
Planning does not by itself show that $U$ is a better latent action.
Let $d_{\mathrm{cos}}(A,B)=1-\operatorname{mean}_{i}\operatorname{cos}(a_i,b_i)$ be the token-averaged cosine distance, where $\operatorname{cos}$ is the cosine similarity of the corresponding token vectors, and let $\operatorname{dist}$ denote this distance.
Write $d_{\mathrm{true}}$ for the prediction error of $F$ under the matched posterior $U$,
$d_{\mathrm{shuffle}}$ when that $U$ is replaced by one from another transition,
and $d_{\mathrm{persist}}=\operatorname{dist}(Z_t,Z_{t+\tau})$.
Action specificity is
\begin{equation}
\nas
=
(d_{\mathrm{shuffle}}-d_{\mathrm{true}})
/
d_{\mathrm{persist}}.
\end{equation}
A larger value means the matched $U$ is more necessary for predicting the future than a shuffled $U$.
Prediction gain on the same distances is
\begin{equation}
\mathrm{PG}
=
(d_{\mathrm{persist}}-d_{\mathrm{true}})
/
d_{\mathrm{persist}}.
\end{equation}

\paragraph{Transition information versus scene information.}
The readouts below ask whether the explicit relation $R$ behaves as a description of what changed.
We freeze each representation and fit linear probes on 4{,}800 transitions from the in-domain navigation training mix, one transition per clip.
Every representation is reduced by averaging over its token axis before the probe, then standardized with the training-split mean and standard deviation.
Paired $\Omega$ features average the $1024$ spatial tokens of each view and concatenate the two views.
The relation is averaged over its nodes, so a structured relation $R\in\mathbb R^{17\times 256}$ is read as a $256$-dimensional vector rather than compared, unpooled, with the $1024\times 2048$ token grid.
Independent feature differences use the same mean pooling on the two visual states.
Motion readout is ridge regression with penalty $1$.
Scene classification uses a linear classifier, the same standardization, and a split that holds out half the transitions of each trajectory.
One probe reads out the physical effect $(\Delta x,\Delta y,\Delta\theta)$.
The other classifies the source trajectory.
Table~\ref{tab:relation_content} compares paired VGGT-$\Omega$ features, the difference of independently encoded visual features, and the relation $R$ in Eq.~\eqref{eq:relation}.
Paired features retain the source trajectory, at $17.7\%$ accuracy, but do not linearly predict the physical effect.
The explicit relation reverses that pattern: motion $R^{2}$ is $0.20$, while trajectory accuracy falls to $1.7\%$.
Independent feature differences predict neither.
The scene probe keeps 297 trajectories that have at least four clips, and a uniform guess among them is correct on $1/297$ of the test cases, about $0.34\%$.
The relation's $1.7\%$ accuracy is above this baseline and far below the paired features, so the drop measures reduced linear identifiability of the source trajectory.

\paragraph{Same-effect retrieval.}
We retrieve nearest neighbors from other trajectories.
A neighbor counts as the same effect when its normalized $(\Delta x,\Delta y,\Delta\theta)$ is closer than $90\%$ of cross-trajectory pairs.
Table~\ref{tab:relation_retrieval} reports the fraction of queries for which at least one of the top-$k$ neighbors meets that criterion.
\method{} places similar motions nearer than either control.

\begin{table}[t]
\caption{
\textbf{Cross-trajectory retrieval of similar effects.}
Neighbors are taken from trajectories other than the query.
}
\label{tab:relation_retrieval}
\centering
\small
\setlength{\tabcolsep}{8pt}
\begin{tabular}{@{}lcc@{}}
\toprule
Representation & Effect Recall@1$\uparrow$ & Effect Recall@5$\uparrow$ \\
\midrule
Paired $\Omega$ features & 22.1 & 60.1 \\
Independent $\Delta Z$ & 14.8 & 48.8 \\
\ourrow
\textbf{\method{} $R$} & \textbf{23.7} & \textbf{60.9} \\
\bottomrule
\end{tabular}
\end{table}

\section{Latent-Objective Ablations}
\label{app:latent_ablate}

Each matched-budget ablation follows the in-domain navigation recipe: training from scratch to 32{,}000 updates, followed by full-state continuation to 40{,}000 updates, as for the full method.
The removed ingredient is the ranking loss, stochastic sampling of $U$, closed-loop rollout, or the KL term.
The deterministic-$U$ ablation uses the posterior mean while retaining the Gaussian head and KL penalty; the other three ablations each remove only the named loss or training rollout.
PG and $\nas$ use the same one-shot RECON protocol as the relation ablation.
The full \method{} result is PG $0.58$, $\nas$ $0.38$, ATE $0.90$, and RPE $0.31$.
Navigation uses the same official RECON evaluation set of 100 trajectories, three proposal seeds, and top-5 execution rule as the full method, with the latent-dependent downstream components retrained for each ablation and checkpoints selected on development data only.
The standalone geometry initialization, which depends only on frozen visual and raw $\Omega$ features, is shared across variants.
All reported values use two decimal places.
Each removal raises ATE and RPE in Table~\ref{tab:latent_ablate}.
Dropping the ranking loss leaves PG at $0.57$ but cuts $\nas$ from $0.38$ to $0.25$, with ATE $1.10$.
A deterministic $U$ and training without closed-loop rollout barely change PG and $\nas$, while ATE rises to $1.03$ and $1.14$.
Removing the KL term raises PG to $0.60$ and $\nas$ to $0.40$, and ATE is still $1.06$.

\begin{table}[t]
\caption{
\textbf{Latent-Objective Ablations.}
Higher is better for PG and $\nas$.
Lower ATE and RPE indicate better performance.
Each row removes one ingredient from the same training and evaluation protocol as the full method.
}
\label{tab:latent_ablate}
\centering
\small
\setlength{\tabcolsep}{4pt}
\begin{tabular}{@{}lcccc@{}}
\toprule
Variant
& \cellcolor{black!8}PG$\uparrow$
& \cellcolor{black!8}$\nas\uparrow$
& \cellcolor{red!10}ATE$\downarrow$
& \cellcolor{red!10}RPE$\downarrow$ \\
\midrule
No ranking loss
& 0.57 & 0.25 & 1.10 & 0.34 \\
Deterministic $U$
& 0.56 & 0.34 & 1.03 & 0.33 \\
No closed-loop rollout
& 0.58 & 0.37 & 1.14 & 0.35 \\
No KL
& 0.60 & 0.40 & 1.06 & 0.34 \\
\ourrow
\textbf{\method{}}
& \textbf{0.58} & \textbf{0.38} & \textbf{0.90} & \textbf{0.31} \\
\bottomrule
\end{tabular}
\end{table}

\section{Structured Relation Modeling}
\label{app:relation_encoder}

This appendix specifies the relation encoder used in Sec.~\ref{sec:relation}.
Shared queries, directed edges, bilinear interactions, and the image-center reference node are design choices for organizing relational transition evidence, rather than the definition of the relation-first principle.
The paired-view tokens, slot extractor, and graph aggregator below are the implemented configuration.

\subsection{Source-Conditioned Slot Extraction}

$\Psi_{\Omega}$ produces $X,Y\in\mathbb R^{1024\times 2048}$, a $32\times 32$ grid.
There are $K_s=16$ scene slots, each of dimension $d_s=256$.
Patch coordinates $c\in\mathbb R^{1024\times 2}$ are cell centers in $[-1,1]$, in row-major order.
The $4\times 4$ anchors use coordinates $\{-0.75,-0.25,0.25,0.75\}$ on each axis, also row-major.
The spatial bias is
\begin{equation}
B_{ij}
=
-\frac{\lVert c_j-a_i\rVert_2^2}{2\sigma^2},
\qquad
\sigma=0.5,
\end{equation}
and the same $B$ is used for the initial readout and for both final readouts.
Attention has four heads, so $d_h=64$.
Content logits are scaled by a gain $g=8$; the bias $B$ is added afterwards and is not multiplied by $g$.
For one head,
\begin{equation}
\begin{aligned}
\Gamma
&=
\operatorname{softmax}\!\left(
g\frac{\widetilde Q W_Q(XW_K)^{\top}}{\sqrt{d_h}}+B
\right),\\
S
&=
W_O\big(\Gamma(XW_V)\big),
\end{aligned}
\label{eq:slot_attention}
\end{equation}
with the softmax over patches.
Head outputs are concatenated before $W_O$.
The initial readout and both final readouts share $W_Q$, $W_K$, $W_V$, and $W_O$.
The query projection has no bias.
Positions $p^{\nu}$ use the attention averaged over heads.
Query conditioning uses $\alpha_q=1$.
$\operatorname{LN}$ is applied on the $256$-dimensional feature, and $\operatorname{Mean}_{K_s}$ subtracts the mean across the $16$ slots.
The two final readouts then share $\widetilde Q_t$ and $B$, one on $X$ and one on $Y$.

\subsection{Directed Edge Encoding and Graph Aggregation}

For each ordered pair of nodes $(i,j)$, the edge at observation $\nu$ is Eq.~\eqref{eq:relation_edge}.
The maps in that equation are bias-free.
$W_d$, $W_q$, and $W_k$ send $256$ dimensions to $128$.
$W_p$ sends the $2$-dimensional image-plane offset to $128$.
$W_b$ maps the $128$-dimensional product to $128$.
The bilinear scale is $\kappa=0.3$.
Features entering $W_q$ and $W_k$ are layer-normalized over the feature dimension.
The image-plane reference node has index $0$, a learned $256$-dimensional feature shared by both observations, and a fixed position $(0,0)$ in the same centered image coordinates.
It is not a camera pose or a robot state.
Together with the $16$ slots this gives $17$ nodes.
Self-loops are removed, leaving $17\times 16=272$ directed edges.
Each edge change is layer-normalized, concatenated with the source-frame offset $p_j^{t}-p_i^{t}$, and scored by a two-layer map with hidden width $64$ and a GELU.
Outgoing weights are a softmax over destinations, and incoming weights are a softmax over sources.
The two weighted sums of $\Delta e$, each in $\mathbb R^{128}$, are concatenated and projected by a bias-free map to one $256$-dimensional relation token.
There is no residual branch and no dropout on this token.
The result is $R_{t,t+\tau}\in\mathbb R^{17\times 256}$.

\subsection{Implementation Diagnostics}

The checks below are initialization measurements on frozen paired-view tokens.
They are not the navigation benchmark, and they explain the adopted structure.

On identical views, $X=Y$, the shared query makes the slot features, image-plane positions, edge changes, and relation tokens agree exactly: the maximum absolute change is $0$.
An asymmetric readout does not have this property, which is why both observations use one query and one bias.

Without the spatial prior, source-view attention maps coincide, with cosine $1.00$, and every patch remains active.
With $\sigma=0.5$ and content gain $8$, the mean source-view attention cosine falls to $0.42$, and the effective support $\exp(H)$ is $553$ of $1024$ patches.
The prior separates the slots at initialization. Changing $Y$ while holding $X$ fixed still moves the future readout, so the content term is not ignored.

At $\kappa=0.3$, the root-mean-square of the scaled bilinear edge change is $0.78$ times the root-mean-square of the linear edge change.
The bilinear term is therefore comparable to the linear term without dominating $\Delta e$.
Setting $\kappa=0$ removes its gradient.

\end{document}